\documentclass[a4paper, 11pt, reqno]{amsart}
\usepackage[utf8]{inputenc}
\usepackage[T1]{fontenc}
\usepackage[top = 1in, bottom = 1in, left = 1in, right = 1in]{geometry}
\usepackage[bottom, hang, flushmargin]{footmisc} 
	\newlength\tindent
	\newcommand\affiliation[2][*]{\begingroup
		\renewcommand\thefootnote{}\footnote{#1 #2}%
		\addtocounter{footnote}{-1}%
		\endgroup}

\usepackage{amsmath}
\usepackage{amsthm}
\usepackage{amsfonts}
\usepackage{bm}
\usepackage{mathtools}
\usepackage{dsfont} 

\usepackage{algorithmic}
\usepackage{etoolbox}
\usepackage[ruled, vlined, linesnumbered]{algorithm2e}
	\DontPrintSemicolon{}
	\SetNlSty{texttt}{}{.}
	\SetAlCapHSkip{0ex}
	\SetAlgoNlRelativeSize{-1}
\usepackage{graphicx}
\usepackage{caption}
	
\usepackage{float} 
\usepackage{tikz}
	\usetikzlibrary{shapes, decorations, arrows, arrows.meta, positioning}
	\tikzset{
		-Latex,
		every edge/.append style = {shorten >= 4pt, shorten <= 4pt},
		state/.style = {ellipse, draw, minimum width = 0.7 cm}
	}

\usepackage{xcolor}
	\definecolor{plotgreen}{RGB}{34, 139, 34}
	\definecolor{plotred}{RGB}{205, 92, 92}
	\definecolor{plotgray}{RGB}{105, 105, 105}

\usepackage{lipsum}
\usepackage{enumitem}
\setlist[itemize]{leftmargin = 0.75cm}
\usepackage{silence}

\usepackage[colorlinks = true, linkcolor = gray, citecolor = gray, urlcolor = gray]{hyperref}
\usepackage[backref = true, style = alphabetic, maxcitenames = 2, maxbibnames = 99, backend = biber]{biblatex}
\newcommand{\citet}[2][]{\citeauthor{#2} \cite[#1]{#2}}
\newcommand{\citep}[2][]{\cite[#1]{#2}}
\usepackage{doi}
\usepackage{cleveref}
	\crefname{equation}{eq.}{eqs.}
	\crefname{line}{line}{lines}

\theoremstyle{plain}
	\newtheorem{theorem}{Theorem}[section]
	\newtheorem{lemma}[theorem]{Lemma}
	\newtheorem{corollary}[theorem]{Corollary}
	\newtheorem{proposition}[theorem]{Proposition}
	
\theoremstyle{remark}
	\newtheorem{remark}[theorem]{Remark}
\theoremstyle{definition}
	\newtheorem{definition}[theorem]{Definition}

\makeatletter
	\def\@tocline#1#2#3#4#5#6#7{\relax
		\ifnum #1>\c@tocdepth 
		\else
			\par \addpenalty\@secpenalty\addvspace{#2}%
			\begingroup \hyphenpenalty\@M
			\@ifempty{#4}{%
				\@tempdima\csname r@tocindent\number#1\endcsname\relax
			}{%
				\@tempdima#4\relax
			}%
			\parindent\z@ \leftskip#3\relax \advance\leftskip\@tempdima\relax
			\rightskip\@pnumwidth plus4em \parfillskip-\@pnumwidth
			#5\leavevmode\hskip-\@tempdima
				\ifcase #1
				\or\or \hskip 1em \or \hskip 2em \else \hskip 3em \fi%
				#6\nobreak\relax
			\dotfill\hbox to\@pnumwidth{\@tocpagenum{#7}}\par
			\nobreak
			\endgroup
		\fi}
\makeatother
\newcommand{\prob}[1]{\mathbb{P}\left(#1\right)}
\newcommand{\mean}[2][\kern-0.5pt]{\mathbb{E}_{#1}\left[#2\right]}
\newcommand{\given}{\, | \,}
\newcommand{\indic}[1]{\mathds{1}_{#1}}
\newcommand{\var}[1]{\operatorname{Var}\left(#1\right)}

\newcommand{\normal}[2]{\mathcal{N}\left(#1, #2\right)}
\newcommand{\invgamma}[2]{\mathcal{IG}\left(#1, #2\right)}
\newcommand{\tstud}[3]{t_{#1}\left(#2, #3\right)}
\newcommand{\wass}[3][2]{\mathcal{W}_{#1}\left(#2, #3\right)}

\newcommand{\real}{\mathbb{R}}

\renewcommand{\natural}{\mathbb{N}}
\newcommand{\borel}[1]{\mathcal{B}\left(#1\right)}
\newcommand{\norm}[1]{\left\lVert#1\right\rVert}
\newcommand{\lipschitz}[1]{\operatorname{Lip}\left(#1\right)}
\renewcommand{\exp}[1]{\operatorname{exp}\left(#1\right)}
\renewcommand{\det}[1]{\operatorname{det}\left(#1\right)}
\newcommand{\trace}[1]{\operatorname{tr}\left(#1\right)}
\newcommand{\flatten}[1]{\operatorname{flat}\left(#1\right)}

    \let\epsilon\varepsilon
    
    \let\oldforall\forall
    \renewcommand{\forall}{\oldforall\,}
    \let\oldexists\exists
    \renewcommand{\exists}{\oldexists\,}

\newcommand{\blankeq}{\kern14.65pt}

\newcommand{\train}{\mathcal{D}}
\newcommand{\test}{\mathcal{T}}
\newcommand{\sigmay}{\sigma_{\scalebox{0.65}{$\kern-3pt\bm{y}_{\mathcal{D}}$}}}

\title[Student-\texorpdfstring{$t$}{t} processes as infinite-width limits of posterior BNNs]{Student-\texorpdfstring{$t$}{t} processes as infinite-width limits of posterior Bayesian neural networks}
\author[F. Caporali]{Francesco Caporali$^*$}
\author[S. Favaro]{Stefano Favaro$^\dag$}
\author[D. Trevisan]{Dario Trevisan$^\ddag$}
\newcommand{\makeaffiliations}{
    \affiliation[$^*$]{Dept. of Operations Research and Financial Engineering, Princeton University \texttt{fc4978@princeton.edu}}
    \affiliation[$^\dag$]{Dept. of Economics and Statistics, University of Torino and Collegio Carlo Alberto, \texttt{stefano.favaro@unito.it}}
    \affiliation[$^\ddag$]{Dept. of Mathematics, University of Pisa, \texttt{dario.trevisan@unipi.it}}
}

\begin{document}

\begin{abstract}
	The asymptotic properties of Bayesian Neural Networks (BNNs) have been extensively studied, particularly regarding their approximations by Gaussian processes in the infinite-width limit. 
	We extend these results by showing that posterior BNNs can be approximated by Student-$t$ processes, which offer greater flexibility in modeling uncertainty. 
	Specifically, we show that, if the parameters of a BNN follow a Gaussian prior distribution, and the variance of both the last hidden layer and the Gaussian likelihood function follows an Inverse-Gamma prior distribution, then the resulting posterior BNN converges to a Student-$t$ process in the infinite-width limit. 
	Our proof leverages the Wasserstein metric to establish control over the convergence rate of the Student-$t$ process approximation.
\end{abstract}

\maketitle
\makeaffiliations

\section{Introduction} \label{sec:intro}

Bayesian neural networks (BNNs), composed of multiple layers of interconnected neurons, have become a powerful tool in modern machine learning, enabling the modeling of complex data structures while quantifying predictive uncertainty \citep{neal1996}. Unlike neural networks (NNs), BNNs offer a solid probabilistic framework where model parameters are treated as random variables with associated probability distributions. In particular, such a framework allows for the incorporation of both prior knowledge and observed data through the prior distribution and likelihood function, respectively.

\subsection{Background and motivation}

The theoretical study of BNNs dates back to the foundational work of \citet{neal1996}, which, inspired by Bayesian principles, showed that wide shallow BNNs converge to Gaussian processes if initialized with independent Gaussian parameters. 
This result was later extended to deep BNNs  \citep{matthews2018,lee2018,basteri2022,favaro2024} as well as to alternative architectures \citep{novak2020,yang2021}, strengthening the connection between deep learning and Gaussian processes in machine learning \citep{gp2006}.

Building on this foundation, significant effort has been devoted to analyzing the posterior behavior of BNNs. Notably, several studies have examined their exact infinite-width limiting posterior distribution, establishing its asymptotic convergence to a Gaussian process \citep{hron2020,trevisan2023}. Parallel research has explored approximate posterior inference methods, including Variational Inference (VI) \citep{blundell2015} and Monte Carlo Markov Chain (MCMC) sampling \citep{izmailov2021,pezzetti2024}, providing an empirical validation of these theoretical results.

Despite the significant progresses in the study of posterior BNNs, existing work typically assumes a fixed variance for the Gaussian prior on the network parameters, a simplification that limits substantially the diversity of posterior behaviors that BNNs can capture. In this paper, we address this critical  limitation by introducing a more flexible model in which the variance itself follows a prior distribution.

\subsection{Our contribution}

Our main contribution is to show that relaxing the fixed-variance assumption in BNNs by using an Inverse-Gamma prior leads to a novel limiting behavior, while preserving the classic prior introduced by \citet{neal1996}. 
Specifically, we prove that while the prior distribution of a BNN converges to a Gaussian process in the infinite-width limit, the marginal posterior distribution converges to a Student-$t$ process. 
The relevance of this result is twofold. First, it provides a new representation of Student-$t$ processes, which have been widely studied and applied in machine learning and statistics \citep{shah2014,tracey2018,sellier2023}. 
Second, it suggests that modifying the prior distributions can yield a broader class of limiting elliptical processes \citep{fang1989,bankestad2020}, opening new research directions on the asymptotic behavior of posterior BNNs. 
This insight shows that a careful selection of prior distributions can enhance model flexibility and uncertainty quantification, offering practical benefits in Bayesian deep learning.

Our approach relies on optimal transport tools, specifically the Wasserstein metric, in order to establish convergence rates and gain control over the distances of the distributions under analysis. Building on prior works \citep{basteri2022,trevisan2023}, we extend the framework to a hierarchical Gaussian-Inverse-Gamma model. In this model, while the prior and likelihood are still assumed to follow multivariate Gaussian distributions with diagonal covariance, the variance of both the last hidden layer and the likelihood function is modeled using an Inverse-Gamma distribution.

\subsection{Outline}

In \Cref{sec:bnns}, we introduce the notation and key tools used in the proof of our main result. 
\Cref{sec:mainresult} presents the main result of the paper, while \Cref{sec:simulations} reports experimental results that serve as a sanity check for the developed theory.

\section{Preliminaries} \label{sec:bnns}

To clarify the discussion, we refer the reader to \Cref{subsec:tensors,subsec:rv,subsec:wassdist} for a review of tensors, random variables, and optimal transport tools.

\subsection{Wasserstein distance}

For a finite set $S$, denote by $\mu$ and $\nu$ two probability measures defined on $(\real^S, \norm{\cdot})$ with finite moment of order $p$, for some $p \geq 1$. The $p$-Wasserstein distance between $\mu$ and $\nu$ is defined as
\begin{equation*}
    \wass[p]{\mu}{\nu} \coloneqq \inf\left\{\mean{\norm{\bm{x} - \bm{y}}^p}^{1/p} \given \bm{x}, \bm{y} \text{ r.v.s with } \mathbb{P}_{\bm{x}} = \mu, \mathbb{P}_{\bm{y}} = \nu\right\},
\end{equation*}
where the infimum is taken over all the random variables $(\bm{x}, \bm{y})$, jointly defined on a probability space $(\Omega, \mathcal{A}, \mathbb{P})$, with marginal laws $\mu$ and $\nu$. 
The random variable $(\bm{x}, \bm{y})$ is referred to as a coupling of $(\mu, \nu)$ and its law, $\gamma$, is called transport plan. We introduce the following abuse of notation: if $\bm{x} \sim \mu$ and $\bm{y} \sim \nu$, $\wass[p]{\bm{x}}{\bm{y}} \coloneqq \wass[p]{\mu}{\nu}$. 

\begin{theorem}[Theorem 6.9 of \citet{otvillani2008}] \label{thm:wassdist}
    Given $(\bm{x}_n)_{n = 1}^{\infty}$, $\bm{x}$ random variables, then $\lim_{n \to \infty} \wass[p]{\bm{x}_n}{\bm{x}} = 0$ if and only if $\bm{x}_n \xrightarrow{law} \bm{x}$, and $\lim_{n \to \infty} \mean{\norm{\bm{x}_n}^p} = \mean{\norm{\bm{x}}^p}$.
\end{theorem}

\subsection{BNNs}

Consider a supervised learning framework in a regression setting, with a given training dataset $\train \coloneqq \left\{\left(\bm{x}_{\mathcal{D}, i}, \bm{y}_{\mathcal{D}, i}\right)\right\}_{i = 1}^k$, i.e.,
\begin{equation*}
    \bm{x}_{\mathcal{D}} \coloneqq \sum_{i = 1}^k \bm{x}_{\mathcal{D}, i} \otimes \bm{e}_i \in \real^{d_{\mathrm{in}} \times k} \quad \text{and} \quad
    \bm{y}_{\mathcal{D}} \coloneqq \sum_{i = 1}^k \bm{y}_{\mathcal{D}, i} \otimes \bm{e}_i \in \real^{d_{\mathrm{out}} \times k}.
\end{equation*}

\begin{definition}[Fully connected feed-forward NN] \label{def:nn}
    A fully connected feed-forward NN is defined through an architecture $\bm{\alpha} \coloneqq (\bm{n}, \bm{\varphi})$ with:
    \begin{enumerate}
        \item $\bm{n}$ denoting the sizes of the $L + 1$ layers\footnote{An input layer, $L - 1$ hidden layers and an output layer.} (with $L \geq 2$)
        \begin{equation*}
            \bm{n} \coloneqq \big(n_0 (\eqcolon d_{\mathrm{in}}), n_1, \dots, n_{L - 1}, n_L (\eqcolon d_{\mathrm{out}})\big), \, n_l \in \natural_{> 0}, \, \forall l = 0, \dots, L;
        \end{equation*}		
        \item $\bm{\varphi}$ denoting the $L$ activation functions (applied component-wise)
        \begin{equation*}
            \bm{\varphi} \coloneqq (\varphi_1, \dots, \varphi_L), \, \varphi_l: \real \to \real, \, \forall l \in [L], \text{ with } \varphi_1(x) = x, \forall x \in \real.
        \end{equation*}
    \end{enumerate}
    In particular, $\forall \bm{x}_0 \in \real^{d_{\mathrm{in}}}$, the NN is defined as $f(\bm{x}_0) \coloneqq f^{(L)}(\bm{x}_0)$ with
    \begin{equation} \label{eq:nn}
        \begin{aligned}
            & f^{(l)}: \real^{d_{\mathrm{in}}} \to \real^{n_l}, \ \forall l = 0, \dots, L, \\
            & f^{(0)}(\bm{x}_0) = \bm{x}_0, \quad f^{(l)}(\bm{x}_0) = \bm{W}^{(l)} \varphi_l\left(f^{(l - 1)}(\bm{x}_0)\right) + \bm{b}^{(l)} \text{ for } l \in [L], \\
        \end{aligned}
    \end{equation}
    where, for any $l \in [L]$, $\bm{W}^{(l)} \in \real^{n_l \times n_{l - 1}}$ and $\bm{b}^{(l)} \in \real^{n_l}$ denote weight matrices and bias vectors, respectively.
\end{definition}

BNNs exploit the power of the Bayes' rule within this supervised learning framework. 
By defining $\bm{\theta} \in \real^t$ (with $t = \sum_{l = 1}^L n_l (n_{l - 1} + 1)$) the flattened concatenation of all parameters of the NN (both weights and biases), it is possible to apply Bayes' theorem to describe the posterior distribution of a BNN:
\begin{equation*}
    \begin{aligned}
        p_{\bm{\theta} | \train}(\bm{\theta}) & = \frac{p_{\train | \bm{\theta}}(\train) \, p_{\bm{\theta}}(\bm{\theta})}{p_{\train}(\train)} 
        \propto  p_{\bm{x}_{\mathcal{D}}, \bm{y}_{\mathcal{D}} | \bm{\theta}} (\bm{x}_{\mathcal{D}}, \bm{y}_{\mathcal{D}}) \, p_{\bm{\theta}}(\bm{\theta}) 
        \propto p_{\bm{y}_{\mathcal{D}} | \bm{\theta}, \bm{x}_{\mathcal{D}}}(\bm{y}_{\mathcal{D}}) \, p_{\bm{\theta}}(\bm{\theta}),
    \end{aligned}
\end{equation*}
where we assumed that $\bm{x}_{\mathcal{D}}$ is independent of $\bm{\theta}$ and all the random variables admit a density with respect to the Lebesgue measure. In particular, if $\mathcal{L}(\bm{\theta}; \bm{y}_{\mathcal{D}}) \coloneqq p_{\bm{y}_{\mathcal{D}} | \bm{\theta}, \bm{x}_{\mathcal{D}}}(\bm{y}_{\mathcal{D}})$ is the likelihood function then
\begin{equation} \label{eq:bayesbnn}
    p_{\bm{\theta} | \train}(\bm{\theta}) \propto \mathcal{L}(\bm{\theta}; \bm{y}_{\mathcal{D}}) \, p_{\bm{\theta}}(\bm{\theta}).
\end{equation}

\begin{definition}[BNN] \label{def:bnn}
    Let $f$ be a NN (see \cref{eq:nn}) with architecture $\bm{\alpha}$. In order to define a BNN we have to put a prior distribution over $\bm{\theta}$ and a likelihood function $\mathcal{L}(\bm{\theta}; \bm{y}_{\mathcal{D}})$ for $\bm{\theta}$ associated to the training set $\train$.
\end{definition}

\begin{remark}
    Bayes' theorem, and a related notion of posterior measure, can be naturally built without the necessity of density functions.
    Let $\bm{\theta}: (\Omega, \mathcal{A}, \mathbb{P}) \to \real^S$ random variable, $S$ finite set, and an evidence $\train$ with density $\mathcal{L}(\bm{\theta}; \bm{y}_{\mathcal{D}})$, we define the posterior measure of $\bm{\theta}$ as
    \begin{equation} \label{eq:bayesbnnmeasure}
        \mathbb{P}_{\bm{\theta} | \train} \coloneqq \frac{\mathcal{L}(\cdot; \bm{y}_{\mathcal{D}})}{\int_{\real^S} \mathcal{L}(\bm{\theta}; \bm{y}_{\mathcal{D}}) d\mathbb{P}_{\bm{\theta}}(\bm{\theta})} \mathbb{P}_{\bm{\theta}},
    \end{equation}
    where, given $\nu$ measure on $(\real^S, \borel{\real^S})$ ($S$ finite set), and $f: \real^S \to \real$, we use the notation $f \nu$ to denote the measure on $(\real^S, \borel{\real^S})$ absolutely continuous with respect to $\nu$ ($f \nu \ll \nu$), with density $f$, i.e., $\forall A \in \borel{\real^S}$, $f \nu (A) = \int_{A} f(\bm{u}) d\nu(\bm{u})$. For the sake of simplicity we always work with densities (as in \cref{eq:bayesbnn}) if they are available, and we swap to the measure theoretic definition in \cref{eq:bayesbnnmeasure} otherwise.
\end{remark}

The prior distribution considered on the parameters is the Gaussian independent prior \citep{neal1996}. Specifically, given the vector of variances $\bm{\sigma} \coloneqq \smash{\big(\big(\sigma_{\bm{W}^{(l)}}^2, \sigma_{\bm{b}^{(l)}}^2\big)\big)_{l = 1}^{L}} \in \left(\real^+ \times \real^+\right)^L$, we assume that
\begin{equation} \label{eq:nealprior}
    \begin{gathered}
        \bm{W}^{(l)} \sim \normal{\bm{0}_{n_l \times n_{l - 1}}}{\sigma_{\bm{W}^{(l)}}^2 / n_{l - 1} \, \bm{I}_{n_l \times n_{l - 1}}}, \ 
        \bm{b}^{(l)} \sim \normal{\bm{0}_{n_l}}{\sigma_{\bm{b}^{(l)}}^2 \bm{I}_{n_l}}, \text{ for } l \in [L].
    \end{gathered}
\end{equation}

The focus of this paper is to study the distribution that the posterior measure of $\bm{\theta}$ induces on $f$, which requires to investigate the behavior of the induced prior. In particular, we need to retrace the well-known results which state that the asymptotic distribution of $f_{\bm{\theta}}(\bm{x}) \coloneqq \sum_{i = 1}^m f_{\bm{\theta}}(\bm{x}_i) \otimes \bm{e}_i$ converges to the neural network Gaussian process (NNGP), where $\bm{x} = \{\bm{x}_i\}_{i = 1}^m$, $m \in \natural_{> 0}$, is a generic input set.

\begin{remark} \label{rem:nncompact}
    In order to have a simpler description of the subsequent theory it is convenient to write the layers of the BNNs in a compact form: $f^{(0)}_{\bm{\theta}}: \real^{d_{\mathrm{in}} \times m} \to \real^{d_{\mathrm{in}} \times m}$, $f^{(0)}_{\bm{\theta}}(\bm{x}) = \bm{x}$ and for every $l \in [L]$, $f^{(l)}_{\bm{\theta}}: \real^{n_{l - 1} \times m} \to \real^{n_l \times m}$,
    \begin{align*}
        f^{(l)}_{\bm{\theta}}(\bm{x}) & = \sum_{i = 1}^m f^{(l)}_{\bm{\theta}}(\bm{x}_i) \otimes \bm{e}_i = \sum_{i = 1}^m \bm{W}^{(l)} \varphi_l\left(f^{(l - 1)}_{\bm{\theta}}(\bm{x}_i)\right) \otimes \bm{e}_i + \bm{b}^{(l)} \otimes \bm{1}_{m} = \\
        & = \left(\bm{W}^{(l)} \otimes \bm{I}_{m}\right) \varphi_l\left(f^{(l - 1)}_{\bm{\theta}}(\bm{x})\right) + \bm{b}^{(l)} \otimes \bm{1}_{m},
    \end{align*}
    where $\bm{W}^{(l)} \otimes \bm{I}_{m} \in \real^{(n_l \times n_{l - 1}) \times (m \times m)}$ should be thought as an element of $\real^{(n_l \times m) \times (n_{l - 1} \times m)}$. We define $f_{\bm{\theta}}(\bm{x}) \coloneqq f^{(L)}_{\bm{\theta}}(\bm{x})$.
\end{remark}

\subsection{NNGP} 

\begin{definition} \label{def:gp}
    Let $H = (H(\bm{x}_0))_{\bm{x}_0 \in \real^{d_{\mathrm{in}}}}$ be a stochastic process such that for any $\bm{x}_0 \in \real^{d_{\mathrm{in}}}$, $H(\bm{x}_0)$ is a random vector with values in $\left(\real^{d_{\mathrm{out}}}, \borel{\real^{d_{\mathrm{out}}}}\right)$. 
    Then $H$ is said to be a Gaussian process with mean function $\bm{M}: \real^{d_{\mathrm{in}}} \to \real^{d_{\mathrm{out}}}$ and covariance kernel $\bm{H}: \real^{d_{\mathrm{in}}} \times \real^{d_{\mathrm{in}}} \to \real^{d_{\mathrm{out}} \times d_{\mathrm{out}}}$, if for any $m > 0$, given $\bm{x} = (\bm{x}_i)_{i = 1}^m \in \real^{d_{\mathrm{in}} \times m}$,
    \begin{equation*}
        H(\bm{x}) \coloneqq \left(H(\bm{x}_1), \dots, H(\bm{x}_m)\right) \sim \normal{\bm{M}(\bm{x})}{\bm{H}(\bm{x})},
    \end{equation*}
    where
    \begin{equation*}
        \bm{M}(\bm{x}) = \left(\bm{M}(\bm{x}_1), \dots, \bm{M}(\bm{x}_m)\right) \in \real^{d_{\mathrm{out}} \times m},
    \end{equation*}
    and $\bm{H}(\bm{x}) \in \real^{(d_{\mathrm{out}} \times m) \times (d_{\mathrm{out}} \times m)}$ can be viewed as a block matrix with $m \times m$ blocks such that, for any $i, j \in [m]^2$, the block $(i, j)$ of $\bm{H}(\bm{x})$ is $\left(\bm{H}(\bm{x})\right)_{i, j} = \bm{H}(\bm{x}_i, \bm{x}_j) \in \real^{d_{\mathrm{out}} \times d_{\mathrm{out}}}$. 
    For such $H$ we set $H \sim \mathcal{GP}\left(\bm{M}, \bm{H}\right)$. 
\end{definition}

\begin{remark}
    In analogy with \Cref{def:gp} it is possible to define Student-$t$ processes replacing the condition on the distribution of $H(\bm{x})$ with a multivariate Student-$t$ with $a$ degrees of freedom: $H(\bm{x}) \sim \tstud{a}{\bm{M}(\bm{x})}{\bm{H}(\bm{x})}$, $a > 0$.
\end{remark}

Following \citet{matthews2018} and \citet{lee2018}, we report below the laws of the random matrices $(G^{(l)}(\bm{x}))_{l = 1}^L$ associated with the infinite-width limits of the $L$ hidden layers of a BNN, evaluated on the input set $\bm{x}$, deriving general expressions for their corresponding Gaussian processes, $(G^{(l)})_{l = 1}^L$:
\begin{equation} \label{eq:nngpx}	
    \begin{aligned}
        & G^{(0)}(\bm{x}) \coloneqq \bm{x} \in \real^{d_{\mathrm{in}} \times m} \text{ constant r.v.}, \\
        & G^{(l)}(\bm{x}) \sim \normal{\bm{0}_{n_l \times m}}{\bm{I}_{n_l} \otimes \bm{K}^{(l)}(\bm{x})}, \text{ with } \bm{K}^{(l)}(\bm{x}) \coloneqq \left(\bm{K}^{(l)}(\bm{x}_i, \bm{x}_j)\right)_{i, j \in [m] \times [m]},
    \end{aligned}
\end{equation}
and, $\forall \bm{x}_0, \bm{x}'_0 \in \real^{d_{\mathrm{in}}}$, $\forall l = 2, \dots, L$, 
\begin{equation*}
    \begin{aligned}
        & \bm{K}^{(1)}(\bm{x}_0, \bm{x}'_0) \coloneqq \sigma_{\bm{W}^{(1)}}^2 (\bm{x}_0^T \bm{x}'_0) / d_{\mathrm{in}} + \sigma_{\bm{b}^{(1)}}^2, \\ 
        & \bm{K}^{(l)}(\bm{x}_0, \bm{x}'_0) \coloneqq \sigma_{\bm{W}^{(l)}}^2 \mean{\varphi_{l}\left(G^{(l - 1)}(\bm{x}_0)_1\right) \varphi_{l}\left(G^{(l - 1)}(\bm{x}'_0)_1\right)} + \sigma_{\bm{b}^{(l)}}^2.
    \end{aligned}
\end{equation*}
Note that, $\forall l \in [L]$, $\bm{I}_{n_l} \otimes \bm{K}^{(l)}(\bm{x}) \in \real^{(n_l \times n_l) \times (m \times m)}$ should be thought reshaped, as elements of $\real^{(n_l \times m) \times (n_l \times m)}$. From \cref{eq:nngpx} it is possible to define the asymptotic Gaussian processes of each hidden layer, $l \in [L]$, as
\begin{equation} \label{eq:nngp}
    G^{(l)} = (G^{(l)}(\bm{x}_0))_{\bm{x}_0 \in \real^{d_{\mathrm{in}}}} \quad \text{and} \quad G^{(l)} \sim \mathcal{GP}\left(\bm{0}, \bm{I}_{n_l} \otimes \bm{K}^{(l)}\right).
\end{equation}
In analogy with the notation introduced for $f_{\bm{\theta}}$ we define $G(\bm{x}) \coloneqq G^{(L)}(\bm{x})$ and $\bm{K}(\bm{x}) \coloneqq \bm{K}^{(L)}(\bm{x})$. We refer to $G = (G(\bm{x}_0))_{\bm{x}_0 \in \real^{d_{\mathrm{in}}}}$ as the NNGP associated to a BNN with architecture $\bm{\alpha}$ and vector of variances $\bm{\sigma}$.

\subsection{Quantitative CLT for prior BNNs} 

\begin{theorem}[\citet{basteri2022, trevisan2023}] \label{thm:priornngp}
    Let $f_{\bm{\theta}}$ BNN, with architecture $\bm{\alpha} = (\bm{n}, \bm{\varphi})$, $\bm{\varphi}$ collection of Lipschitz activation functions, a prior on $\bm{\theta}$ as in \cref{eq:nealprior}, $\left(G^{(l)}\right)_{l = 1}^L$ Gaussian processes as in \cref{eq:nngp} and an input set $\bm{x} \in \real^{d_{\mathrm{in} \times m}}$. Then, $\forall l \in [L]$ exists a constant $c > 0$ independent of $\left(n_j\right)_{j = 1}^l$, such that,
    \begin{equation} \label{eq:priornngp}
        \wass[p]{f_{\bm{\theta}}^{(l)}(\bm{x})}{G^{(l)}(\bm{x})} \leq c \sqrt{n_l} \sum_{j = 1}^{l - 1} \frac{1}{\sqrt{n_j}}.
    \end{equation}
\end{theorem}

The constant $c$ in \cref{eq:priornngp} in general depends on the input set $\bm{x}$, that must be finite. We remark that quantitative functional bounds, i.e., for infinitely many inputs, have been also established by \citet{favaro2024}.

\begin{remark} \label{rem:gaussposterior}
    An immediate consequence of \Cref{thm:priornngp} (achievable applying \Cref{thm:wassdist}) is that, by letting $n_l$ grow to $\infty$, the process $f^{(l)}_{\bm{\theta}}$ associated with the $l$-th hidden layer of a BNN evaluated on $\bm{x}$ converges in distribution to the NNGP's component $G^{(l)}(\bm{x})$: given $n_{min} \coloneqq \min_{j = 1, \dots, l - 1} n_j$,
    \begin{equation*}
        f^{(l)}_{\bm{\theta}}(\bm{x}) \xrightarrow[n_{min} \to \infty]{law} \normal{\bm{0}_{n_l \times m}}{\bm{I}_{n_l} \otimes \bm{K}^{(l)}(\bm{x})}.
    \end{equation*} 
    Therefore, \Cref{thm:priornngp} yields a quantitative version of what has already been proved by \citet{matthews2018,lee2018}.	
\end{remark}

\section{\texorpdfstring{Student-$t$}{Student-t} approximation of posterior BNNs} \label{sec:mainresult}

Our goal is to extend the closeness result between the \textit{induced} prior distribution on the BNN, $f_{\bm{\theta}}$, and the corresponding NNGP, $G$, established in \Cref{thm:priornngp}, to their respective \textit{induced} posterior distributions. In particular, the main result of this section, \Cref{thm:studposterior}, provides a posterior counterpart of \Cref{thm:priornngp}.

\subsection{Law of posterior NNGP} \label{subsec:nngplaw}

It is useful to start by introducing the hierarchical model applied to the NNGP. In particular, by assuming
\begin{equation} \label{eq:hiervariancenngp}
    \begin{gathered}	
        \sigma_{\bm{W}^{(l)}}^2, \sigma_{\bm{b}^{(l)}}^2 \text{ constants, } \forall l = 1, \dots, L - 1, \\
        \sigma^2 \coloneqq \sigma_{\bm{W}^{(L)}}^2 = \sigma_{\bm{b}^{(L)}}^2 = \sigmay^2 \quad \text{and} \quad \sigma^2 \sim \invgamma{a}{b}, \text{ with } a, b > 0,
    \end{gathered}
\end{equation}
we have
\begin{equation} \label{eq:hiermodel}
    \begin{aligned}
        G(\bm{x}_{\mathcal{D}}) \given \sigma^2 & \sim \normal{\bm{0}_{n_L \times k}}{\sigma^2 \bm{I}_{n_L} \otimes \bm{K}'(\bm{x}_{\mathcal{D}})}, \\
        \sigma^2 & \sim \invgamma{a}{b}, \text{ Inverse-Gamma with } a, b > 0, \\
        \bm{y}_{\mathcal{D}} \given G(\bm{x}_{\mathcal{D}}), \sigma^2 & \sim \normal{G(\bm{x}_{\mathcal{D}})}{\sigma^2 \bm{I}_{n_L \times k}},
    \end{aligned}
\end{equation}
with
\begin{align} \label{eq:rescaledkernel}
    \bm{K}'(\bm{x}_{\mathcal{D}}) = \mean{\varphi_{L}\left(G^{(L - 1)}(\bm{x}_{\mathcal{D}})\right)^T \varphi_{L}\left(G^{(L - 1)}(\bm{x}_{\mathcal{D}})\right)} / n_L + \bm{1}_{k \times k},
\end{align}
rescaled NNGP kernel independent of $\sigma^2$. Therefore, observing that
\begin{equation*}
    p_{G(\bm{x}_{\mathcal{D}}), \sigma^2 | \train}(\bm{z}, \sigma^2) \propto p_{\bm{y}_{\mathcal{D}} | G(\bm{x}_{\mathcal{D}}), \sigma^2}(\bm{y}_{\mathcal{D}}) \, p_{G(\bm{x}_{\mathcal{D}}) | \sigma^2}(\bm{z}) \, p_{\sigma^2}(\sigma^2),
\end{equation*}
assuming $\bm{K}'(\bm{x}_{\mathcal{D}})$ to be invertible, $n_L = 1$, and applying standard tricks (see \Cref{sec:postnngp}), we obtain that
\begin{equation*}
    \begin{aligned}
        G(\bm{x}_{\mathcal{D}}) \given \sigma^2, \train & \sim \normal{\bm{y}_{\mathcal{D}} \bm{M}^{-1}}{\sigma^2 \bm{M}^{-1}}, \\
        \sigma^{2} \given \train & \sim \invgamma{a + \frac{k}{2}}{b + \frac{1}{2} \left(\bm{y}_{\mathcal{D}} \left(\bm{I}_{k} - \bm{M}^{-1}\right) (\bm{y}_{\mathcal{D}})^T\right)}.
    \end{aligned}
\end{equation*}
Hence,
\begin{equation*}
    \begin{aligned}
        G(\bm{x}_{\mathcal{D}}) \given \train \sim \tstud{2a + k}{\bm{\mu}_{\mathrm{post}}}{\bm{\Sigma}_{\mathrm{post}}},
    \end{aligned}
\end{equation*}
with
\begin{equation} \label{eq:tstudpostparams}
    \begin{aligned}
        \bm{M} & \coloneqq \bm{I}_{k} + \bm{K}'(\bm{x}_{\mathcal{D}})^{-1}, \\
        \bm{\mu}_{\mathrm{post}} & \coloneqq \bm{y}_{\mathcal{D}} \bm{M}^{-1}, \\
        \bm{\Sigma}_{\mathrm{post}} & \coloneqq \left(b + \frac{1}{2} \left(\bm{y}_{\mathcal{D}} \left(\bm{I}_{k} - \bm{M}^{-1}\right) (\bm{y}_{\mathcal{D}})^T\right)\right) \frac{2}{2a + k} \bm{M}^{-1}.
    \end{aligned}
\end{equation}

\begin{remark} \label{rem:poststudtprocess}
    In a completely analogous way, it is possible to show that, given an input test set $\bm{x}_{\mathcal{T}} \in \real^{n_0 \times k'}$, 
    \begin{equation} \label{eq:poststudtprocess}
        \begin{aligned}
            & G(\bm{x}_{\mathcal{T}}) \given \train \sim \tstud{2a + k}{\bm{\mu}'_{\mathrm{post}}}{\left(b + \frac{1}{2} \left(\bm{y}_{\mathcal{D}} \left(\bm{I}_{k} - \bm{M}^{-1}\right) (\bm{y}_{\mathcal{D}})^T\right)\right) \frac{2}{2a + k} \bm{\Sigma}'_{\mathrm{post}}}, \text{ with } \\
            & \kern20pt \bm{\mu}'_{\mathrm{post}} \coloneqq \bm{K}'(\bm{x}_{\mathcal{T}}, \bm{x}_{\mathcal{D}}) \left(\bm{K}'(\bm{x}_{\mathcal{D}}) + \sigma^2 I_{k}\right)^{-1} \bm{y}_{\mathcal{D}}, \\
            & \kern20pt \bm{\Sigma}'_{\mathrm{post}} \coloneqq \bm{K}'(\bm{x}_{\mathcal{T}}) - \bm{K}'(\bm{x}_{\mathcal{T}}, \bm{x}_{\mathcal{D}}) \left(\bm{K}'(\bm{x}_{\mathcal{D}}) + \sigma^2 I_{k}\right)^{-1} \bm{K}'(\bm{x}_{\mathcal{T}}, \bm{x}_{\mathcal{D}}). \\
        \end{aligned}
    \end{equation}
    Indeed, following the strategy adopted by \citet[eqs. (2.22) to (2.24)]{gp2006}, we can observe that
    \begin{align*}
        & G(\bm{x}_{\mathcal{T}}) \given \sigma^2, \train \sim \normal{\bm{\mu}'_{\mathrm{post}}}{\bm{\Sigma}'_{\mathrm{post}}}, \\
        & \sigma^2 \given \train \sim \invgamma{a + \frac{k}{2}}{b + \frac{1}{2} \left(\bm{y}_{\mathcal{D}} \left(\bm{I}_{k} - \bm{M}^{-1}\right) (\bm{y}_{\mathcal{D}})^T\right)},
    \end{align*}
    which in turn implies \cref{eq:poststudtprocess} by means of \Cref{lem:norminvgammastudent}.
\end{remark}

\subsection{Posterior BNNs} \label{subsec:postbnn}

We define $\widetilde{\mu} \sim f_{\bm{\theta}}(\bm{x})$, $\mu \sim G(\bm{x})$, with $\bm{x} = (\bm{x}_{\mathcal{D}}, \bm{x}_{\mathcal{T}}) \in \real^{n_0 \times (k + k')}$, fixed input set which extends the input training set with a possible input test set, and omit the dependence on $\bm{y}_{\mathcal{D}}$ in the Gaussian likelihood $\mathcal{L}$\footnote{Now $\mathcal{L}$ depends also on $s \coloneqq \sigma^2$, which is no more a parameter.}, where
\begin{equation} \label{eq:likelihood}
    \mathcal{L}: \real^{n_L \times k} \times \real^{+} \to \real, \ \mathcal{L}(\bm{z}, s) = \frac{1}{\left(2 \pi s\right)^{n_L k / 2}} \exp{-\frac{1}{2 s} \norm{\bm{y}_{\mathcal{D}} - \bm{z}}_F^2}.
\end{equation}
The objective is to bound the $1$-Wasserstein distance between the marginal posterior of the BNN, $f_{\bm{\theta}}(\bm{x})$, and the marginal posterior of the NNGP evaluated on the input set, $G(\bm{x})$. The latter can be found integrating with respect to $s$ the prior measures $\mu$ and $\widetilde{\mu}$, both multiplied by the prior density of the variance $p_{\sigma^2}(s)$ and the likelihood $\mathcal{L}(\cdot, s)$.
In formulas, we aim to find an upper bound for $\wass[1]{\widetilde{\mu}_{\mathrm{post}}}{\mu_{\mathrm{post}}}$, with $\mu_{\mathrm{post}} \sim G(\bm{x}) \given \train$, $\widetilde{\mu}_{\mathrm{post}} \sim f_{\bm{\theta}}(\bm{x}) \given \train$, probability measures defined as in the following \Cref{def:postnngpbnn}.

\begin{remark}
    The likelihood function can can be extended to the space $\real^{n_L \times (k + k')}$, by artificially making it depend on the test input set while disregarding its contribution. 
    Consequently, the entire reasoning developed below extends naturally to this more general case through a straightforward change of variables. 
    However, to maintain a simpler notation and ensure a coherent presentation, we state and prove our main result under the framework introduced in \Cref{subsec:nngplaw}, i.e., assuming $\bm{x} = \bm{x}_{\mathcal{D}}$.
\end{remark}

\begin{definition} \label{def:postnngpbnn}
    Given a BNN as in \Cref{def:bnn}, we assume to have a hierarchical model as the one described in \cref{eq:hiermodel} for the NNGP, and an analogous model for the BNN (i.e., Gaussian prior on $\bm{\theta}$ as in \cref{eq:nealprior}, prior on the variance $\sigma^2$ as in \cref{eq:hiervariancenngp} and a likelihood as in \cref{eq:likelihood}). Then, for any $A \in \borel{\real^{n_L \times k}}$, we define the posterior measures as follows: given $\mu \sim G(\bm{x})$ and $\widetilde{\mu} \sim f_{\bm{\theta}}(\bm{x})$,
    \begin{equation*}
        \begin{aligned}
            & \mu_{\mathrm{post}}(A) \coloneqq \int_A \int_{\real^+} \frac{1}{\mathit{I}} \mathcal{L}(\bm{z}, s) p_{\sigma^2}(s) ds \mu(d\bm{z}), \ \mathit{I} \coloneqq \int_{\real^{n_L \times k}} \int_{\real^+} \mathcal{L}(\bm{z}, s) p_{\sigma^2}(s) ds \mu(d\bm{z}), \\
            & \widetilde{\mu}_{\mathrm{post}}(A) \coloneqq \int_A \int_{\real^+} \frac{1}{\widetilde{\mathit{I}}} \mathcal{L}(\bm{z}, s) p_{\sigma^2}(s) ds \widetilde{\mu}(d\bm{z}), \ \widetilde{\mathit{I}} \coloneqq \int_{\real^{n_L \times k}} \int_{\real^+} \mathcal{L}(\bm{z}, s) p_{\sigma^2}(s) ds \widetilde{\mu}(d\bm{z}).
        \end{aligned}
    \end{equation*}
\end{definition}

\subsection{Main result}

Building on \Cref{def:postnngpbnn}, the main result of this work can be summarized in the following \Cref{thm:studposterior}.

\begin{theorem} \label{thm:studposterior}
    Let $f_{\bm{\theta}}$, $G$, $\bm{x}$ and $\bm{y}_{\mathcal{D}}$ as above, $\mathcal{L}$ density of a $\normal{\bm{z}}{\sigmay^2 \bm{I}_{n_L \times k}}$. Assume a common variance for the last hidden layer of the BNN and the likelihood, distributed as an Inverse-Gamma 
    \begin{equation*}
        \sigma^2 \coloneqq \sigma_{\bm{W}^{(L)}}^2 = \sigma_{\bm{b}^{(L)}}^2 = \sigmay^2, \ \sigma^2 \sim \invgamma{a}{b},
    \end{equation*}
    with
    \begin{equation} \label{eq:abconstr}
        a > \frac{1}{2}, \ b > \left(1 + \frac{\epsilon + 2}{2\epsilon + 2}\right) \norm{\bm{y}_{\mathcal{D}}}_F^2, \text{ for any } \epsilon < 1/\norm{\bm{K}'(\bm{x})}_{\mathrm{op}}. 
    \end{equation}
    Then, there exists a constant $c > 0$, independent of $\left(n_l\right)_{l = 1}^{L - 1}$, such that
    \begin{equation*}
        \wass[1]{f_{\bm{\theta}}(\bm{x}) \given \train}{G(\bm{x}) \given \train} \leq \frac{c}{\sqrt{n_{min}}}.
    \end{equation*}
\end{theorem}
\begin{proof}[Sketch of the proof] 
    The idea is to show the convergence of $f_{\bm{\theta}}(\bm{x}) \given \train$ to $G(\bm{x}) \given \train$ through the following steps: 
    \begin{enumerate}
        \item partially retracing the strategy introduced by \citet{trevisan2023}, we first prove that there exist some constants $c > 0$ independent of $\left(n_l\right)_{l = 1}^{L - 1}$ and $\sigma^2$, and a function $h: \real^+ \to \real^+$ independent on $\left(n_l\right)_{l = 1}^{L - 1}$ as well, such that
        \begin{equation} \label{eq:simpleconvposteriorsigma}
            \wass[1]{f_{\bm{\theta}}(\bm{x}) \given (\train, \sigma^2)}{G(\bm{x}) \given (\train, \sigma^2)} \leq h(\sigma^2) \frac{c_0}{\sqrt{n_{min}}};
        \end{equation}
        \item we try to apply the convexity property of $1$-Wasserstein distance in the following \Cref{prop:wassconvexity} (\hyperlink{proof:wassconvexity}{proof} in \Cref{subsec:wassdist}) to the families of probabilities $\left(\mathbb{P}_{f_{\bm{\theta}}(\bm{x}) \given (\train, \sigma^2)}\right)_{\sigma^2 \in \real^+}$ and $\left(\mathbb{P}_{G(\bm{x}) \given (\train, \sigma^2)}\right)_{\sigma^2 \in \real^+}$, which would lead to the thesis.
        \begin{proposition} \label{prop:wassconvexity}
            Let us consider two Markov kernels $(\mu(s))_{s \in \real^+}$, $(\widetilde{\mu}(s))_{s \in \real^+}$ with source $\real^+$ and target $\real^T$ and a measure $\nu$ on $\real^+$ ($T$ finite set).
            Defining the probability measures on $\real^T$ such that, $\forall B \in \borel{\real^T}$,
            \begin{equation*}
                \mu(B) \coloneqq \int_{\real^+} \mu(s)(B) d\nu(s), \quad \widetilde{\mu}(B) \coloneqq \int_{\real^+} \widetilde{\mu}(s)(B) d\nu(s),
            \end{equation*}
            the following convexity property for the distance $\mathcal{W}_1$ holds: 
            \begin{equation} \label{eq:wassconvex}
                \wass[1]{\mu}{\widetilde{\mu}} \leq \int_{\real^+} \wass[1]{\mu(s)}{\widetilde{\mu}(s)} d\nu(s).
            \end{equation}
        \end{proposition}
    \end{enumerate}
    Unfortunately the second step is not easy as it could seem since \Cref{prop:wassconvexity} requires two families of probability measures (in particular Markov kernels) integrated with respect to the same measure $\nu$, which is not exactly our setting. 
    Let us describe the issue before approaching the solution. For this purpose it is useful to introduce
    \begin{equation*}
        \begin{aligned}
            \mathit{I}_{\sigma^2}(s) \coloneqq \int_{\real^{n_L \times k}} \mathcal{L}(\bm{z}, s) \mu(d\bm{z}), \quad \widetilde{\mathit{I}}_{\sigma^2}(s) \coloneqq \int_{\real^{n_L \times k}} \mathcal{L}(\bm{z}, s) \widetilde{\mu}(d\bm{z}),
        \end{aligned}
    \end{equation*}
    which allow us to write, $\forall A \in \borel{\real^{n_L \times k}}$,
    \begin{equation} \label{eq:mupost}
        \begin{aligned}
            & \mu_{\mathrm{post}}(A) = \int_{\real^+} \mu_{\sigma^2}(s)(A) \frac{\mathit{I}_{\sigma^2}(s)}{\mathit{I}} p_{\sigma^2}(s) ds, \text{ with } \mu_{\sigma^2}(s)(A) \coloneqq \int_A \frac{\mathcal{L}(\bm{z}, s) \mu(d\bm{z})}{\mathit{I}_{\sigma^2}(s)}, \\
            & \widetilde{\mu}_{\mathrm{post}}(A) = \int_{\real^+} \widetilde{\mu}_{\sigma^2}(s)(A) \frac{\widetilde{\mathit{I}}_{\sigma^2}(s)}{\widetilde{\mathit{I}}} p_{\sigma^2}(s) ds, \text{ with } \widetilde{\mu}_{\sigma^2}(s)(A) \coloneqq \int_A \frac{\mathcal{L}(\bm{z}, s) \widetilde{\mu}(d\bm{z})}{\widetilde{\mathit{I}}_{\sigma^2}(s)}.
        \end{aligned}
    \end{equation}
    Eventually, we can note that the two families $(\mu_{\sigma^2}(s))_{s \in \real^+}$ and $(\widetilde{\mu}_{\sigma^2}(s))_{s \in \real^S}$, which coincide respectively with $\left(\mathbb{P}_{G(\bm{x}) \given (\train, \sigma^2)}\right)_{\sigma^2 \in \real^+}$ and $\left(\mathbb{P}_{f_{\bm{\theta}}(\bm{x}) \given (\train, \sigma^2)}\right)_{\sigma^2 \in \real^+}$, are integrated with respect to different measures.
    Hence, it is clear that to apply \cref{eq:wassconvex} it is necessary to use the triangle inequality,
    \begin{equation} \label{eq:trianstep}
        \wass[1]{\mu_{\mathrm{post}}}{\widetilde{\mu}_{\mathrm{post}}} \leq \wass[1]{\mu_{\mathrm{post}}}{\bar{\mu}} + \wass[1]{\bar{\mu}}{\widetilde{\mu}_{\mathrm{post}}},
    \end{equation}
    where we inserted a third measure,
    \begin{equation} \label{eq:barmu}
        \bar{\mu}(A) \coloneqq \int_{\real^+} \widetilde{\mu}_{\sigma^2}(s)(A) \frac{\mathit{I}_{\sigma^2}(s)}{\mathit{I}} p_{\sigma^2}(s) ds, \ \forall A \in \borel{\real^{n_L \times k}},
    \end{equation}
    specifically constructed to satisfy the hypothesis of the convexity property. \\
    Now, to conclude, we just need to control both the terms on the right-hand side of \cref{eq:trianstep}.
    \begin{itemize}[leftmargin = 2.5cm]
        \item[\textit{1\textsuperscript{st} term.}] We just apply the aforementioned convexity property getting
        \begin{equation*}
            \wass[1]{\mu_{\mathrm{post}}}{\bar{\mu}} \leq \int_{\real^+} \wass[1]{\mu_{\sigma^2}(s)}{\widetilde{\mu}_{\sigma^2}(s)} \frac{\mathit{I}_{\sigma^2}(s)}{\mathit{I}} p_{\sigma^2}(s) ds.
        \end{equation*}
        Therefore by \cref{eq:simpleconvposteriorsigma} we get 
        \begin{equation*}
            \wass[1]{\mu_{\mathrm{post}}}{\bar{\mu}} \leq \frac{c_0}{\sqrt{n_{min}}} \int_{\real^+} h(s) \frac{\mathit{I}_{\sigma^2}(s)}{\mathit{I}} p_{\sigma^2}(s) ds \leq \frac{c_1}{\sqrt{n_{min}}},
        \end{equation*}
        where, in order to bound the last integral it is necessary to introduce a constraint on $a$ and $b$, as in \cref{eq:abconstr}.
        \item[\textit{2\textsuperscript{nd} term.}] We exploit the following technical \Cref{lem:boundwasstv}, in combination with several bounds on the first moments of the considered probability measures.
        \begin{lemma} \label{lem:boundwasstv}
            Let $\mu$, $\nu$ be measures on $\real^S$, then, denoting with $|\mu - \nu|$ the total variation measure, it holds
            \begin{equation*}
                \wass[1]{\mu}{\nu} \leq \int_{\real^S} \norm{\bm{u}} d |\mu - \nu|(\bm{u}).
            \end{equation*}
        \end{lemma}
    \end{itemize}
    \vskip-10pt
\end{proof}

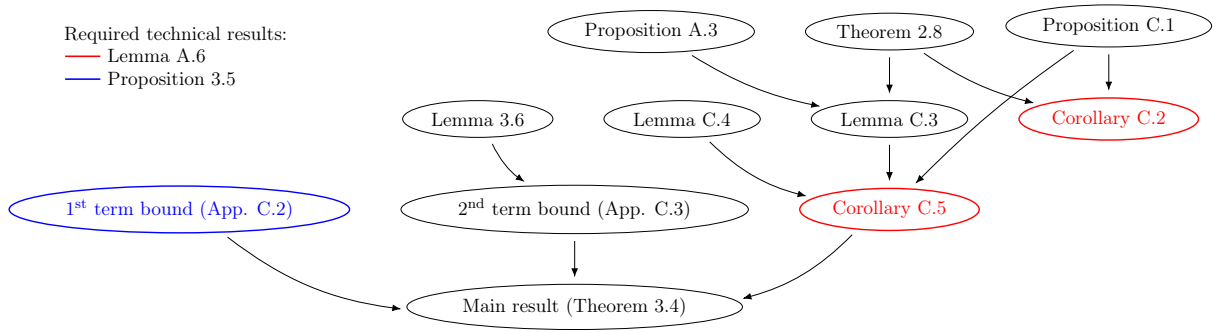
\begin{figure}[H]
	\begin{centering}
		\resizebox{\textwidth}{!}{
		\begin{tikzpicture}
			\hypersetup{colorlinks = true, linkcolor = black}
			\node[state] (main) at (0,0) {Main result (\Cref{thm:studposterior})};
			\node[state] (secondbound) [above = of main] {2\textsuperscript{nd} term bound (App. \ref{subsec:secondterm})};
			\hypersetup{colorlinks = true, linkcolor = blue}
			\node[state] (firstbound) [left = of secondbound, draw = blue, thick] {\textcolor{blue}{1\textsuperscript{st} term bound (App. \ref{subsec:firstterm})}};
			\hypersetup{colorlinks = true, linkcolor = black} 
			\hypersetup{colorlinks = true, linkcolor = red}
			\node[state] (convposteriorsigma) [right = of secondbound, draw = red, thick] {\textcolor{red}{\Cref{cor:convposteriorsigma}}};
			\hypersetup{colorlinks = true, linkcolor = black} 
			\path (firstbound) edge[bend right = 15] (main);
			\path (secondbound) edge[] (main);
			\path (convposteriorsigma) edge[shorten >= 4pt, bend left = 15] (main);
			\node[state] (priornngpsigma) [above = of convposteriorsigma] {\Cref{lem:priornngpsigma}};
			\node[state] (boundexplikelihood) [left = of priornngpsigma] {\Cref{lem:boundexplikelihood}};
			\node[state] (boundwasstv) [left = of boundexplikelihood] {\Cref{lem:boundwasstv}};
			\hypersetup{colorlinks = true, linkcolor = red}
			\node[state] (convposterior) [right = of priornngpsigma, draw = red, thick] {\textcolor{red}{\Cref{cor:convposterior}}};
			\hypersetup{colorlinks = true, linkcolor = black}
			\path (boundwasstv) edge[bend right = 15] (secondbound);
			\path (boundexplikelihood) edge[bend right = 15] (convposteriorsigma);
			\path (priornngpsigma) edge[] (convposteriorsigma);
			\node[state] (priornngp) [above = of priornngpsigma] {\Cref{thm:priornngp}};
			\node[state] (likelihood) [above = of convposterior] {\Cref{prop:likelihood}};
			\node[state] (wassmult) [left = of priornngp] {\Cref{prop:wassmult}};
			\path (wassmult) edge[bend right = 10] (priornngpsigma);
			\path (priornngp) edge[] (priornngpsigma);
			\path (priornngp) edge[bend right = 10] (convposterior);
			\path (likelihood) edge[] (convposterior);
			\path (likelihood) edge[bend right = 5] (convposteriorsigma);
				\node[anchor = north east, text width = 6cm] at (-4, 5.75) {%
					Required technical results: \\
					\raisebox{3pt}{\tikz{\draw[red, solid, line width = 0.9pt, -](0, 0) -- (7mm, 0);}} \Cref{lem:nnposterior} \\ 
					\raisebox{3pt}{\tikz{\draw[blue, solid, line width = 0.9pt, -](0, 0) -- (7mm, 0);}} \Cref{prop:wassconvexity}
				};
		\end{tikzpicture}}
		\caption{Dependency of results for the proof of \Cref{thm:studposterior}.}
		\label{fig:chartmain}
	\end{centering}
\end{figure}

\begin{remark}
    All the details of the sketched part of the previous proof can be found in \Cref{sec:mainproof}: a formal statement and a proof of \cref{eq:simpleconvposteriorsigma} can be found in \Cref{subsec:convposteriorsigma}; additionally, the bounds for the two terms found using the triangle inequality (\cref{eq:trianstep}) can be found respectively in \Cref{subsec:firstterm} and \Cref{subsec:secondterm}. 
    In \Cref{fig:chartmain}, we include a chart illustrating the dependencies among the results that lead to the proof of \Cref{thm:studposterior}.
\end{remark}

By exploiting the connection between $\mathcal{W}_1$ and weak convergence, together with what was observed in \Cref{subsec:nngplaw}, \Cref{thm:studposterior} leads us to a characterization of the asymptotic behavior of the exact posterior law of a BNN trained following the Gaussian-Inverse-Gamma model, showing convergence to a Student-t process in the infinite-width limit.

\begin{corollary} \label{cor:studposterior}
    Under the same assumptions of \Cref{thm:studposterior} and $n_L = 1$, the posterior of the BNN $f_{\bm{\theta}}$ with Gaussian-Inverse-Gamma prior and Gaussian likelihood, evaluated in the input set $\bm{x}$, converges in law to a multivariate Student-$t$ variable with $2a+k$ degrees of freedom:
    \begin{equation*}
        f_{\bm{\theta}}(\bm{x}) \given \train \xrightarrow[n_{min} \to \infty]{law} \tstud{2a + k}{\bm{\mu}_{\mathrm{post}}}{\bm{\Sigma}_{\mathrm{post}}},
    \end{equation*}
    with $M$, $\bm{\mu}_{\mathrm{post}}$ and $\bm{\Sigma}_{\mathrm{post}}$ as in \cref{eq:tstudpostparams}.
\end{corollary}

\section{Simulations} \label{sec:simulations}

We present a procedure to sample from the posterior distribution of a BNN, ensuring consistency between the theoretical results and practical implementations. 
We consider a hierarchical Bayesian model, where we place a prior on both the network parameters, $\bm{\theta} =$ $= (\bm{W}^{(1)}, \bm{b}^{(1)}, \dots, \bm{W}^{(L)}, \bm{b}^{(L)})$, and the variance $\sigma^2$:
\begin{align} \label{eq:bnnhiermodel}
    & \bm{W}^{(l)} \sim \normal{\bm{0}_{n_l \times n_{l - 1}}}{\sigma_{\bm{W}^{(l)}}^2 / n_{l - 1} \, \bm{I}_{n_l \times n_{l - 1}}}, \ && \bm{b}^{(l)} \sim \normal{\bm{0}_{n_l}}{\sigma_{\bm{b}^{(l)}}^2 \bm{I}_{n_l}},  \text{ for } l \in [L - 1], \nonumber \\
    & \bm{W}^{(L)} \given \sigma^2 \sim \normal{\bm{0}_{n_L \times n_{L - 1}}}{\sigma^2 / n_{L - 1} \, \bm{I}_{n_L \times n_{L - 1}}}, \ && \bm{b}^{(L)} \given \sigma^2 \sim \normal{\bm{0}_{n_L}}{\sigma^2 \bm{I}_{n_L}}, \\
    & \sigma^2 \sim \invgamma{a}{b}, \ && \bm{y}_{\mathcal{D}} \given \bm{\theta}, \bm{x}_{\mathcal{D}}, \sigma^2 \sim \normal{f_{\bm{\theta}}(\bm{x}_{\mathcal{D}})}{\sigma^2 \bm{I}_{n_L \times k}}. \nonumber
\end{align}

\begin{algorithm}
    \SetAlgoVlined
    \LinesNumbered
    \caption{\textsc{Training of a BNN under Gaussian-Inverse-Gamma prior}}
    \label{alg:samplingpostbnn}
    \KwIn{$\bm{\alpha}$ [architecture], $\bm{\sigma}$ [BNN's variances], $(a, b)$ [Inverse-Gamma parameters], \newline
        $f$ [function]}
    \textbf{build} training set $\train$ starting from a reference function $f$ \;
    \textbf{build} test set $\test$ based on a fine partitioning of the domain of $f$ \;
    \textbf{initialize} $\smash{\sigma^2_{(0)}}$ \;
    \For{$i = 0, \dots m$}{
        \textbf{sample} $\bm{\theta}_{(i + 1)} \given \sigma^2, \train$ using \textbf{\texttt{NUTS}} \label{line:bnnsample} \;
        \textbf{sample} $\smash{\sigma^2_{(i + 1)}} \given \bm{\theta}_{(i)}, \train$ from $\smash{p_{\sigma^2 | \bm{\theta}_{(i + 1)}, \train}(\sigma^2)}$ \label{line:varsample} \;
    }
    \Return $f_{\bm{\theta}_{(m)}}(\test) \given \train$ \;
\end{algorithm}

The Monte Carlo sampling strategy is summarized in \Cref{alg:samplingpostbnn}. In particular, the idea is to sample from $\bm{\theta}, \sigma^2 \given \train$ using a Gibbs sampling scheme. 
To implement \cref{line:bnnsample} we rely on MCMC methods, which have been widely studied in the literature of Bayesian optimization for BNNs. 
Several versions of such strategies are implemented in the \texttt{Python} library \texttt{Pyro} \citep{pyro,pytorch}, and among these, we applied the No-U-turn sampler \citep{hoffman2011}. 
Whereas, in order to sample from the marginal posterior of the variance (\cref{line:varsample}) we exploit the strategy adopted by \citet[Appendix D]{ding2022}, since it is possible to show that it follows a positive Generalized Inverse Gaussian distribution (referred as $\mathcal{GIN}^+$ by the authors). 
Such a derivation, as well as additional implementation details, can be found in \Cref{sec:simulationsdetails}.

\begin{figure}
    \centering
    \includegraphics[width = \textwidth]{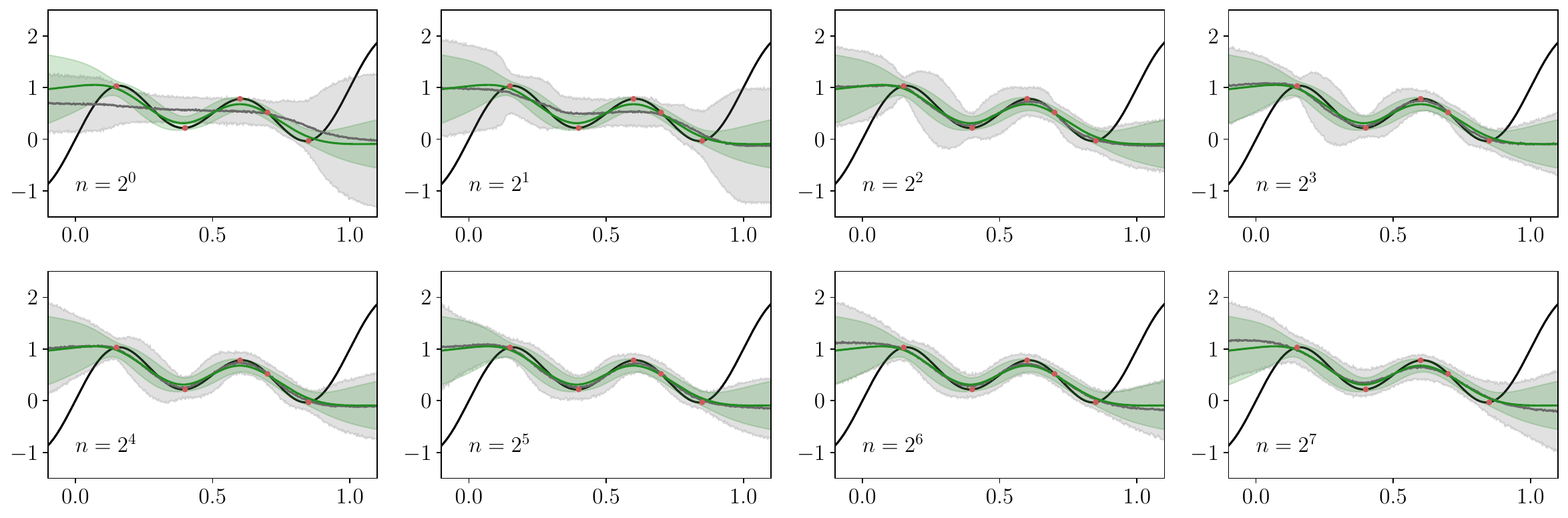}
    \captionsetup{width = .95\textwidth, font = small, skip = 2pt}
    \caption{%
        Sequence of posterior BNNs, \textcolor{plotgray}{$(f_{\bm{\theta}_n} \given \train)_n$} (in gray), converging to the corresponding posterior Student-$t$ process, \textcolor{plotgreen}{$G \given \train$} (in green), in the infinite-width limit. 
        Given \textcolor{plotred}{$\train$} (in red), training set, we sampled $100$ values from both $G \given \train$ and $f_{\bm{\theta}_n} \given \train$ for each width $n \in \{2^0, \dots, 2^7\}$, following \Cref{rem:poststudtprocess} and \Cref{alg:samplingpostbnn}, respectively. 
        The networks used have $2$ hidden layers, \texttt{erf} activations and parameter variances set to $5$. 
        Additionally, the hyperparameters $(a, b)$ are set to $(3, 2)$. 
    }
    \label{fig:normalinvgammaprior}
\end{figure}

\begin{figure}
    \centering
    \includegraphics[width = \textwidth]{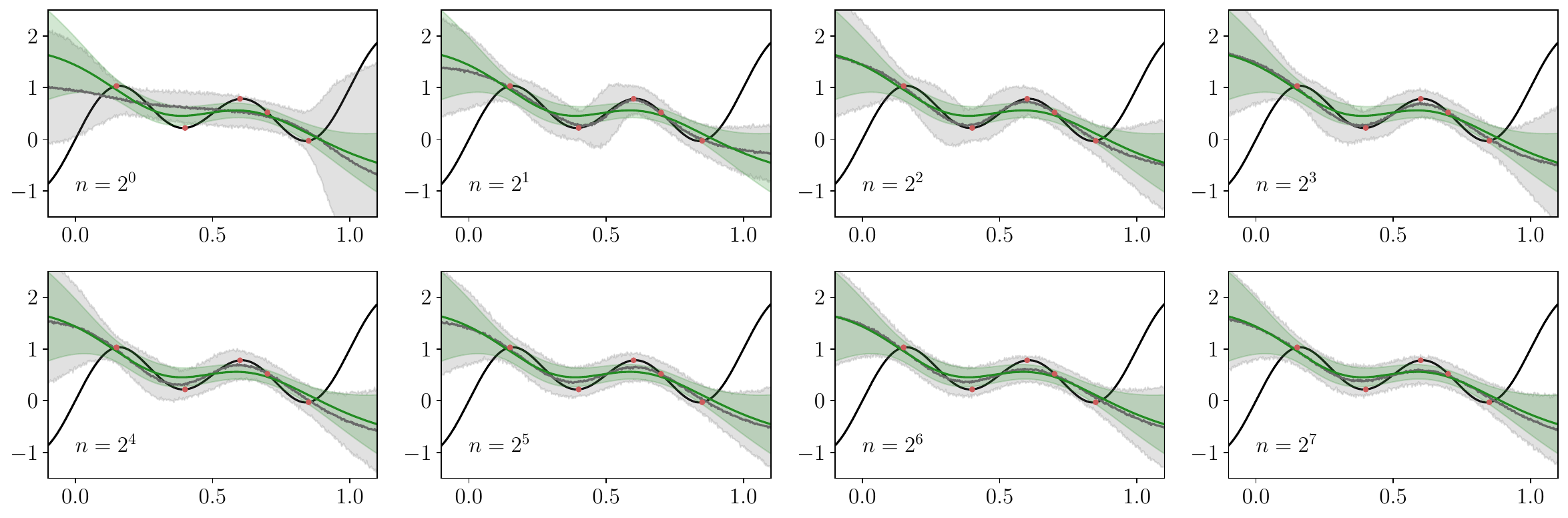}
    \captionsetup{width = .95\textwidth, font = small, skip = 2pt}
    \caption{%
        Sequence of posterior BNNs, \textcolor{plotgray}{$(f_{\bm{\theta}_n} \given \train)_n$} (in gray), converging to the corresponding posterior Gaussian process, \textcolor{plotgreen}{$G \given \train$} (in green), in the infinite-width limit. 
        Given \textcolor{plotred}{$\train$} (in red), training set, we sampled $100$ values from both $G \given \train$ and $f_{\bm{\theta}_n} \given \train$ for each width $n \in \{2^0, \dots, 2^7\}$. 
        The sampling was performed following \citet[eqs. (2.22)-(2.24)]{gp2006} for $G \given \train$ and the built-in \texttt{NUTS} algorithm in \texttt{Pyro} for $f_{\bm{\theta}_n} \given \train$. 
        The networks used have $2$ hidden layers, \texttt{erf} activations, parameter variances set to $2$, and likelihood variance set to $0.1$.
    }
    \label{fig:gaussianprior}
\end{figure}

We report in \Cref{fig:normalinvgammaprior} the comparison between a sequence of BNNs trained using the strategy discussed above and the limiting Student-$t$ process discussed in \Cref{sec:postnngp} (see \Cref{rem:poststudtprocess}). 
As the (all equal) widths of the hidden layers increase, the models' output distributions become closer. 
We also consider the case in which we replace the Gaussian-Inverse-Gamma prior with the classical Gaussian prior with fixed variance (see \Cref{fig:gaussianprior}). 
In this case the limiting process to which the sequence of BNNs converges is simply the posterior NNGP (see \citet{hron2020,trevisan2023}).

\begin{remark}
    We can observe that in \Cref{fig:gaussianprior}, i.e., under Gaussian prior, the convergence is much faster and more precise compared to the Gaussian-Inverse-Gamma prior case (\Cref{fig:normalinvgammaprior}). 
    This behavior, while likely influenced by our specific sampling procedure for the posterior BNNs, is also consistent with theoretical expectations.
    Indeed, although the theoretical convergence rates of the limiting processes are identical for both the Gaussian prior and the Gaussian-Inverse-Gamma prior cases (see \Cref{cor:convposterior} and \Cref{thm:studposterior}), the associated multiplicative constants differ significantly in magnitude. 
    Therefore, given a common fixed width, different distances between the posterior BNNs and their limiting processes are to be expected.
\end{remark}

We conclude by comparing the asymptotic processes under both frameworks.
The posterior Student-$t$ process models the variance of the data more accurately compared to the posterior Gaussian process.
This is an expected behavior, as the Gaussian-Inverse-Gamma model explicitly estimates the data variance during the Bayesian learning, whereas no such estimation is performed when we use a Gaussian prior.
This final result also highlights that using a Gaussian-Inverse-Gamma prior provides a more accurate representation of the data, particularly in scenarios in which the dataset is relatively small.

\begin{figure}
    \centering
    \includegraphics[width = .9\textwidth]{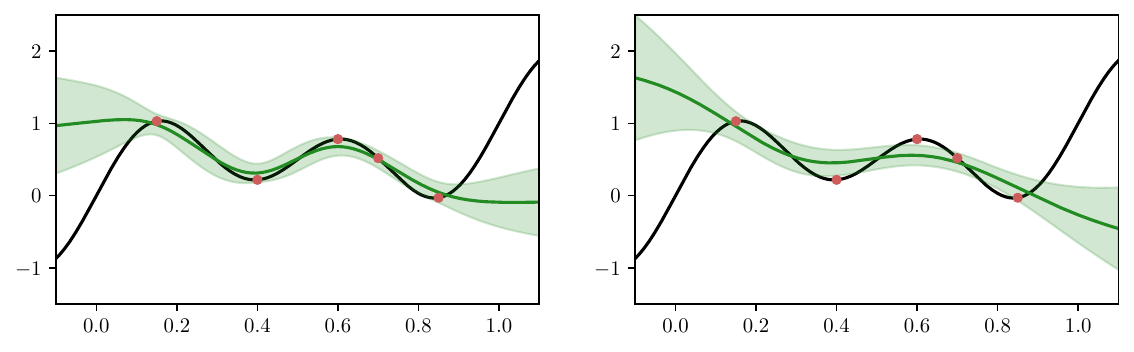}
    \captionsetup{width = .9\textwidth, font = small, skip = 2pt}
    \caption{%
        Posterior Student-$t$ process (on the right) and posterior Gaussian process (on the left). 
        We followed the same strategy and used the same parameters introduced to generate \Cref{fig:normalinvgammaprior,fig:gaussianprior}.
    }
    \label{fig:comparison}
\end{figure}

{\sloppy\printbibliography}
\clearpage

\markboth{\MakeUppercase{\shortauthors}}{\MakeUppercase{\shorttitle}}

\appendix

\hypersetup{colorlinks = true, linkcolor = black}
\tableofcontents
\hypersetup{colorlinks = true, linkcolor = gray}

\section{Notation} \label{sec:notation}

We summarize below the preliminary tools and notations used throughout the article. Proofs of standard results are omitted, with references provided. For clarity, the content is organized into thematic sections.

\subsection{Tensors} \label{subsec:tensors}

Given $S$ finite set (e.g., $[n] \text{ or } [n_1] \times \dots \times [n_k], \text{ for some } k \geq 2$, where $[n] =$ $= \{1, \dots, n\}$) we denote with $\real^S$ the vector space of real valued functions $\bm{v}: S \to \real$ (column vectors which generalizes to multidimensional tensors). 
We adopt the notation $\bm{v}_s \coloneqq \bm{v}(s)$, $\forall s \in S$, and we introduce the following conventions: $\bm{e}_s$ is the $s$\textsuperscript{th} vector of the canonical base ($\bm{e}_s(s) = 1$ and $\bm{e}_s(r) = 0$, $\forall r \in S \setminus \{s\}$), $\bm{1}_{S} \coloneqq \sum_{s \in S} \bm{e}_s$ and $\bm{0}_{S}$ is the constant null vector. 
When $S$ is used as a subscript/superscript we also simplify the notation: $[n_1] \times \dots \times [n_k]$ becomes just $n_1 \times \dots \times n_k$. \\
Given $S$, $T$ two finite sets, we can see the vector space $\real^{S \times T}$ as the space of linear transformations $\bm{A}: \real^T \to \real^S$, defining 
\begin{equation*}
	\forall \bm{v} \in \real^T, \, \bm{A} \bm{v} \coloneqq \bm{A}(\bm{v}) = \sum_{s \in S} \left(\sum_{t \in T} \bm{A}_{s, t} \bm{v}_t\right) \bm{e}_s.
\end{equation*}
If $S = [n]$ and $T = [m]$ then $\bm{A}$ can be represented as a standard matrix $\bm{A} \in \real^{m \times n}$, where $\bm{A}_{i, j} \coloneqq (\bm{A}(\bm{e}_j))(i)$, $\forall i \in [n], j \in [m]$. 
Analogously if $S = [n_1] \times [n_2]$ and $T = [m_1] \times [m_2]$ it is possible to represent $\bm{A}$ as a $4$-dimensional tensor such that $\bm{A}_{i, j, k, l} \coloneqq \bm{A}(\bm{e}_{(k, l)})((i, j))$, $\forall (i, j) \in [n_1] \times [n_2], (k, l) \in [m_1] \times [m_2]$. \\
If $S = T$, we introduce the notation $\operatorname{Sym}^{S}$ for the set of symmetric linear transformations in $\real^{S \times S}$, i.e., $\bm{A} \in \operatorname{Sym}^{S}$ if and only if $\bm{A} \in \real^{S \times S}$, $\bm{A} = \bm{A}^T$. 
Moreover, we denote with $\operatorname{Sym}_+^{S}$ the subset of $\operatorname{Sym}^{S}$ composed by symmetric, positive definite matrices, where a matrix $\bm{A}$ is said to be positive definite if $\forall \bm{v} \in \real^S \setminus \{0\}$, $\bm{v}^T \bm{A} \bm{v} > 0$.

\smallskip

We define the outer product (or tensor product) between $\bm{v} \in \real^S$ and $\bm{w} \in \real^{T}$ as $\bm{v} \otimes \bm{w} \in \real^{S \times T}$ with $(\bm{v} \otimes \bm{w})_{s, t} \coloneqq \bm{v}_s \bm{w}_t$, $\forall s \in S, t \in T$, denoting $\bm{v} \otimes \bm{v}$ as $\bm{v}^{\otimes 2}$. 
We also define the identity map $\bm{I}_S: \real^S \to \real^S$ as $\bm{I}_S \coloneqq \sum_{s \in S} \bm{e}_s^{\otimes 2}$, observing that, $\forall \bm{v} \in \real^S$, $\bm{I}_S \bm{v} = \sum_{s \in S} (\sum_{t \in S} (\bm{I}_S)_{s, t} \bm{v}_t) \bm{e}_s =$ $= \sum_{s \in S} \bm{v}_s \bm{e}_s = \bm{v}$.
If $S = [n]$, $T = [m]$ then $\bm{v} \otimes \bm{w} = \bm{v} \bm{w}^T$. \\
Given a generic pair of elements in a real vector space, $\bm{v}, \bm{w} \in \real^S$, we define the standard scalar product between them as 
\begin{equation*}
	\langle \bm{v}, \bm{w} \rangle \coloneqq \trace{\bm{v} \otimes \bm{w}} = \sum_{s \in S} \bm{v}_s \bm{w}_s.
\end{equation*}
Analogously we can define the Euclidean norm (or $2$-norm) induced by the scalar product as
\begin{equation} \label{eq:tensornorm}
	\norm{\bm{v}} = \langle \bm{v}, \bm{v} \rangle^{1/2} = \left(\sum_{s \in S} \bm{v}_s^2\right)^{1/2}.
\end{equation}
If $S = [n]$, $n \in \natural_{> 0}$, we also define the $p$-norm of $\bm{v}$, $p \geq 1$, as $\norm{\bm{v}}_p \coloneqq \left(\sum_{i = 1}^{n} \bm{v}_i^p\right)^{1/p}$. 
If nothing is specified we always consider $\real^n$ as a normed spaced with the Euclidean norm defined in \cref{eq:tensornorm}. 
In the matrix case, $S = [n] \times [m]$, this norm is usually referred as Frobenius norm, therefore for an improved readability we denote it as $\norm{\cdot}_F$. \\
Given an operator $\bm{A} \in \operatorname{Sym}^{S}$, with $S = [n]$, we define the operator norm as
\begin{equation*}
	\norm{\bm{A}}_{\mathrm{op}} \coloneqq \sup_{\norm{\bm{x}}_2 = 1} \norm{\bm{A}\bm{x}}_2 = \max\left\{\lambda \given \lambda \in \operatorname{Sp}(\bm{A})\right\}.
\end{equation*}

\begin{lemma} \label{lem:norminequalities}
	For every $\bm{x}, \bm{y} \in \left(\real^S, \norm{\cdot}\right)$ real vector space with the Euclidean norm, the following inequalities hold:
	\begin{align}
		\norm{\bm{x} - \bm{y}}^2 & \leq 2 (\norm{\bm{x}}^2 + \norm{\bm{y}}^2), \label{eq:ineqnormsqleq} \\
		\norm{\bm{x} - \bm{y}}^2 & \geq \frac{\epsilon}{\epsilon + 1} \norm{\bm{x}}^2 -\epsilon \norm{\bm{y}}^2, \forall \epsilon \in \real^+ \label{eq:ineqnormsqgeqeps}. 
	\end{align}
\end{lemma}
\begin{proof}
	We prove both the statements using the triangle inequality and its inverse: $\forall \bm{u}, \bm{v} \in \real^S$, $\norm{\bm{u} + \bm{v}} \leq \norm{\bm{u}} + \norm{\bm{v}}$ and $\norm{\bm{u} - \bm{v}} \geq \norm{\bm{u}} - \norm{\bm{v}}$. \\
	Let us start observing that $\forall a, b, \epsilon > 0$, we have
	\begin{equation*}
		\begin{aligned}
			(a + b)^2 & = a^2 + b^2 + \epsilon 2a \frac{b}{\epsilon} \leq a^2 + b^2 + \epsilon \left(a^2 + \frac{b^2}{\epsilon^2}\right) = a^2 \left(1 + \epsilon\right) + b^2 \left(1 + \frac{1}{\epsilon}\right).
		\end{aligned}
	\end{equation*}
	With $\epsilon = 1$ we get $(a + b)^2 \leq 2(a^2 + b^2)$, and therefore by triangle inequality  follows the first result,
	\begin{equation*}
		\norm{\bm{x} - \bm{y}}^2 \leq \left(\norm{\bm{x}} + \norm{\bm{y}}\right)^2 \leq 2 \left(\norm{\bm{x}}^2 + \norm{\bm{y}}^2\right).
	\end{equation*}
	For the second inequality we use the reverse triangle inequality,
	\begin{equation*}
		\begin{aligned}
			\norm{\bm{x}}^2 & \leq \left(\norm{\bm{x} - \bm{y}} + \norm{\bm{y}}\right)^2 \leq \norm{\bm{x} - \bm{y}}^2 (1 + \epsilon) + \norm{\bm{y}} \left(1 + \frac{1}{\epsilon}\right),
		\end{aligned}
	\end{equation*}
	which implies
	\begin{equation*}
		\norm{\bm{x} - \bm{y}}^2 \geq \norm{\bm{x}}^2 \frac{1}{1 + \epsilon} - \norm{\bm{y}}^2 \frac{1}{\epsilon}.
	\end{equation*}
	Now by simply substituting $\epsilon' = \epsilon^{-1}$ in place of $\epsilon$ we get $(1 + (\epsilon')^{-1})^{-1} = \epsilon' / (\epsilon' + 1)$ and so the thesis.
\end{proof}

\subsection{Random variables} \label{subsec:rv}

Given $S$ finite set, a random variable $\bm{x}$ with values in $\real^S$ is a measurable map
\begin{equation*}
	\bm{x}: (\Omega, \mathcal{A}, \mathbb{P}) \to (\real^S, \borel{\real^S}).
\end{equation*}
It is adopted the same notation for deterministic tensors introduced in \Cref{subsec:tensors} and random tensors, i.e., random variable with values in tensor spaces. 
We denote with $\mathbb{P}_{\bm{x}}$ the distribution (or law) of $\bm{x}$, 
\begin{equation*}
	\mathbb{P}_{\bm{x}} \coloneqq \prob{\bm{x}^{-1}(A)}, \, \forall A \in \borel{\real^S}.
\end{equation*}
We write $\bm{x} \sim \bm{y}$ if two random variables share the same distribution.

\smallskip

Given $\bm{x}$ random variable with values on $\real^S$ we define its mean value, or first moment, and its variance\footnote{Often referred as covariance if $|S| > 1$.}, or second moment of $\bm{x} - \mean{\bm{x}}$, respectively as
\begin{equation*}
	\begin{aligned}
		\mean{\bm{x}} & \coloneqq \left(\mean{\bm{x}_s}\right)_{s \in S} = \left(\int_{\Omega} \bm{x}_s(\omega) d\mathbb{P}(\omega)\right)_{s \in S} \in \real^S,
	\end{aligned}
\end{equation*}
and
\begin{equation*}
	\begin{aligned}
		\var{\bm{x}} & \coloneqq \mean{\left(\bm{x} - \mean{\bm{x}}\right)^{\otimes 2}} \in \operatorname{Sym}_+^{S}.
	\end{aligned}
\end{equation*}
Given $p \geq 1$, we define the Lebesgue norm of order $p$ of $\bm{x}$ as
\begin{equation*}
	\norm{\bm{x}}_{L^p} \coloneqq \mean{\norm{\bm{x}}^p}^{1/p} \in \real.
\end{equation*}
Recalling that $\norm{\bm{x}}^2 = \trace{\bm{x}^{\otimes 2}}$, thanks to the linearity of the integral, it is possible to exchange trace and expectation so that we can write 
\begin{equation*}
	\norm{\bm{x}}_{L^2}^2 = \mean{\trace{\bm{x}^{\otimes 2}}} = \trace{\mean{\bm{x}^{\otimes 2}}}.
\end{equation*}

\subsubsection{Gaussian random variables}

Given $\bm{\mu} \in \real^S$, $\bm{\Sigma} \in \operatorname{Sym}_+^{S}$, we denote with $\normal{\bm{\mu}}{\bm{\Sigma}}$ the distribution of a Gaussian random variable $\bm{x}: (\Omega, \mathcal{A}, \mathbb{P}) \to (\real^S, \borel{\real^S})$, such that
\begin{equation*}
	\mean{\bm{x}} = \bm{\mu} \quad \text{and} \quad \var{\bm{x}} = \mean{\left(\bm{x} - \bm{\mu}\right)^{\otimes 2}} = \bm{\Sigma}.
\end{equation*}
\begin{remark} \label{rem:gaussmoments} 
	Let $\left(\real^S, \norm{\cdot}\right)$ be a normed space with a Euclidean norm and $\bm{\mu} = \bm{0}_S$. Then,
	\begin{equation*}
	\mean{\norm{\bm{x}}}^2 \leq \mean{\norm{\bm{x}}^2} = \mean{\sum_{s \in S} \bm{x}_s^2} = \sum_{s \in S} \mean{\bm{x}_s^2} = \trace{\bm{\Sigma}}.
	\end{equation*}
\end{remark}

\subsubsection{Inverse-Gamma random variables}

A random variable $s$, with values in $\left(\real^+, \borel{\real^+}\right)$, is said to be Inverse-Gamma distributed with parameters $a$ and $b$ in $\real^+$ (denoted as $s \sim \invgamma{a}{b}$), if $\mathbb{P}_{s}$ admits a density with respect to the Lebesgue measure $\lambda^+$ on $\real^+$, and in particular
\begin{equation*}
	p_s(s) \coloneqq \frac{d\mathbb{P}_{s}}{d\lambda^+}(s) = \frac{b^a}{\Gamma(a)} \left(\frac{1}{s}\right)^{a + 1} \exp{-\frac{b}{s}}.
\end{equation*}
As the name suggests, and a simple change of variables shows, one can equivalently say that the variable $1 / s$ is Gamma distributed with shape and rate parameters $(a, b)$.

\subsubsection{Multivariate \texorpdfstring{Student-$t$}{Student-t} random variables}

Given $k \in \natural_{> 0}$, a random variable $\bm{z}$, with values in $\left(\real^k, \borel{\real^k}\right)$, is said to be $k$-dimensional Student-$t$ distributed with $\nu$ degrees of freedom, location $\bm{\mu} \in \real^k$ and scale $\bm{\Sigma} \in \operatorname{Sym}_+^{k}$ (denoted as $\bm{z} \sim \tstud{\nu}{\bm{\mu}}{\bm{\Sigma}}$), if $\mathbb{P}_{\bm{z}}$ admits a density with respect to the Lebesgue measure $\lambda^k$ on $\real^k$, and in particular
\begin{equation*}
	p_{\bm{z}}(\bm{z}) \coloneqq \frac{d\mathbb{P}_{\bm{z}}}{d\lambda^k}(\bm{z}) = \frac{\Gamma((\nu + k)/2)}{\Gamma(\nu/2) (\nu \pi)^{k/2} \det{\bm{\Sigma}}^{1/2}} \left(1 + \frac{1}{\nu} (\bm{z} - \bm{\mu})^T \bm{\Sigma}^{-1} (\bm{z} - \bm{\mu})\right)^{-\frac{\nu + k}{2}}.
\end{equation*}

\subsection{Wasserstein distance} \label{subsec:wassdist}

We first recall two well-known properties (see e.g., \citet{otvillani2008}) of the Wasserstein metric, together with the main two technical tools exploited in the demonstration of our core result: \Cref{prop:wassconvexity} and \Cref{lem:nnposterior}.

\begin{proposition} \label{prop:wassmult}
	Given two random variables $\bm{x}$, $\bm{y}$ with values in $\real^S$ and a positive constant $a$, $\forall p \geq 1$, it holds
	\begin{equation} \label{eq:wasslinear}
		\wass[p]{a \bm{x}}{a \bm{y}} = a \wass[p]{\bm{x}}{\bm{y}}.
	\end{equation}
\end{proposition}

\begin{theorem}[Kantorovich duality for $\mathcal{W}_1$] \label{thm:kantorovichduality}
	Given $T$ finite set, $\mu$ and $\widetilde{\mu}$ probability measures on $\real^T$, then we have
	\begin{equation} \label{eq:kantorovichduality}
		\wass[1]{\mu}{\widetilde{\mu}} = \sup_{\substack{f: \real^T \to \real, \\ \lipschitz{f} \leq 1}} \left(\int_{\real^T} f(\bm{x}) d\mu(\bm{x}) - \int_{\real^T} f(\bm{x}) d\widetilde{\mu}(\bm{x})\right). 
	\end{equation}
\end{theorem}

The supremum above is computed over all the functions $f: \real^T \to \real$ that are Lipschitz continuous, with Lipschitz constant $\lipschitz{f} \leq 1$. Notice that one can further restrict to functions $f$ such that $f(0) = 0$ since adding constants to $f$ does not change the difference of the integrals.

\smallskip

Before we state and prove \Cref{prop:wassconvexity}, let us recall the notion of Markov kernel. 

\begin{definition}[Markov kernel]
	Let us consider $(E, \mathcal{E})$, $(F, \mathcal{F})$ measurable spaces, a Markov kernel with source $(E, \mathcal{E})$ and target $(F, \mathcal{F})$ is a map $K_{\mu}: E \times \mathcal{F} \to [0, 1]$, such that
	\begin{itemize}
		\item[-] $\forall B \in \mathcal{F}$, the map $s \to K_{\mu}(s, B)$ for $s \in E$ is measurable from $(E, \mathcal{E})$ to $([0, 1], \borel{[0, 1]})$;
		\item[-] $\forall s \in E$, the map $B \to K_{\mu}(s, B)$ for $B \in \mathcal{F}$ is a probability measure on $(F, \mathcal{F})$.
	\end{itemize}
	For any fixed $s \in E$, we denote $\mu(s) \coloneqq K_{\mu}(s, \cdot)$ and the Markov kernel as $K_{\mu} = (\mu(s))_{s \in E}$.
\end{definition}
\begin{proof}[\normalfont\bfseries Proof of \Cref{prop:wassconvexity}] \hypertarget{proof:wassconvexity}
	Consider $f: \real^T \to \real$ with $\lipschitz{f} \leq 1$. Then by \cref{eq:kantorovichduality} with the measures $\mu(s)$ and $\widetilde{\mu}(s)$ it follows
	\begin{equation} \label{eq:wassconvexproof}
		\int_{\real^T} f(\bm{x}) d\mu(s)(\bm{x}) - \int_{\real^T} f(\bm{x}) d\widetilde{\mu}(s)(\bm{x}) \leq \wass[1]{\mu(s)}{\widetilde{\mu}(s)}.
	\end{equation}
	Integrating both sides in \cref{eq:wassconvexproof} with respect to $\nu$ yields
	\begin{equation*} 
		\int_{\real^T} f(\bm{x}) d\mu(\bm{x}) - \int_{\real^T} f(\bm{x}) d\widetilde{\mu}(\bm{x}) \leq \int_{\real^+} \wass[1]{\mu(s)}{\widetilde{\mu}(s)} d\nu(s).
	\end{equation*}
	We conclude by taking the supremum over the possible $f$'s and again by Kantorovich duality \cref{eq:kantorovichduality}).
\end{proof}

The following \Cref{lem:nnposterior} shows that if two \textit{prior} distributions are close with respect to the Wasserstein metric and the (common) Likelihood is sufficiently regular, then also the \textit{posterior} distributions will be close, in a quantitative way.

\smallskip

We use the notation
\begin{equation*}
	m_{p}(\mu) \coloneqq \int_{\real^S} \norm{\bm{z}}^{p} d \mu(\bm{z}).
\end{equation*}
for the moment of order $p \geq 1$ of a measure $\mu$.

\begin{lemma}[Lemma 5.1 of \citet{trevisan2023}] \label{lem:nnposterior}
	Let $\mu$, $\widetilde{\mu}$ be probability measures on $(\real^S, \norm{\cdot})$ for some finite set $S$ and finite moments of order $p \geq 1$.
	Fix $g: \real^S \to \real^+$ be a uniformly bounded (by $\norm{g}_\infty$) Lipschitz continuous map (with constant $\lipschitz{g}$), such that
	\begin{equation*}
		\mu(g) \coloneqq \int_{\real^S} g(\bm{z}) d\mu(\bm{z}) > 0 \quad \text{and} \quad \widetilde{\mu}(g) >0.
	\end{equation*}
	Defining the probability measures $\mu_g \ll \mu$ and $\widetilde{\mu}_g \ll \widetilde{\mu}$, with respective densities $\frac{d\mu_g}{d\mu} \coloneqq \frac{g}{\mu(g)}$ and $\frac{d\widetilde{\mu}_g}{d\widetilde{\mu}} \coloneqq \frac{g}{\widetilde{\mu}(g)}$, it holds
	\begin{equation} \label{eq:wassposterior}
		\wass[1]{\widetilde{\mu}_g}{\mu_g} \leq \frac{1}{\mu(g)} \left(\lipschitz{g} m_{p / (p - 1)}(\mu) + \left(1 + \frac{m_1(\mu) \lipschitz{g}}{\widetilde{\mu}(g)}\right) \norm{g}_{\infty}\right) \wass[p]{\widetilde{\mu}}{\mu}.
	\end{equation}
\end{lemma}

\section{Posterior NNGP} \label{sec:postnngp}

Given the Bayesian framework presented in \cref{eq:hiervariancenngp,eq:hiermodel,eq:rescaledkernel}, in order to get the posterior distribution of $G(\bm{x}_{\mathcal{D}}) \given \train$ we write explicitly all the densities and apply the Bayes rule. 
In particular, assuming $\bm{K}'(\bm{x}_{\mathcal{D}}) \in \operatorname{Sym}_+^{k}$ invertible (we already know that is symmetric positive semi-definite), flattening all the random matrices by columns and defining
\begin{equation*}
	\bm{y}_{\mathrm{f}} \coloneqq \flatten{\bm{y}_{\mathcal{D}}}, \quad \bm{z}_{\mathrm{f}} \coloneqq \flatten{\bm{z}},
\end{equation*}
with $\bm{z} \coloneqq G(\bm{x}_{\mathcal{D}}) \in \real^{n_L \times k}$, we get
\begin{equation*}
	\begin{aligned}	
		p_{G(\bm{x}_{\mathcal{D}}) | \sigma^2}(\bm{z}) & = \frac{1}{\left((2\pi \sigma^2)^{n_L k} \det{\bm{K}'(\bm{x}_{\mathcal{D}}) \otimes_K \bm{I}_{n_L}}\right)^{1/2}} \cdot \\
		& \blankeq \cdot \exp{-\frac{1}{2\sigma^2} \bm{z}_{\mathrm{f}}^T \left(\bm{K}'(\bm{x}_{\mathcal{D}}) \otimes_K \bm{I}_{n_L}\right)^{-1} \bm{z}_{\mathrm{f}}}, \\
		p_{\sigma^2}(\sigma^2) & = \frac{b^a}{\Gamma(a)} \frac{1}{\left(\sigma^2\right)^{a + 1}} \exp{-\frac{b}{\sigma^2}}, \\
		p_{\bm{y}_{\mathcal{D}} | G(\bm{x}_{\mathcal{D}}), \sigma^2}(\bm{y}_{\mathcal{D}}) &= \frac{1}{\left(2\pi\sigma^2\right)^{n_L k / 2}} \exp{-\frac{1}{2 \sigma^2} (\bm{y}_{\mathrm{f}} - \bm{z}_{\mathrm{f}})^T (\bm{y}_{\mathrm{f}} - \bm{z}_{\mathrm{f}})}.
	\end{aligned}
\end{equation*}
with $\otimes_K$ representing the Kronecker product. \\
By performing explicit computation we retrieve the posterior distribution of $G(\bm{x}_{\mathcal{D}}), \sigma^2 \given \train$ as
\begin{equation*}
	\begin{aligned}
		p_{G(\bm{x}_{\mathcal{D}}), \sigma^2 | \train}(\bm{z}, \sigma^2) & \propto p_{\bm{y}_{\mathcal{D}} | G(\bm{x}_{\mathcal{D}}), \sigma^2}(\bm{y}_{\mathcal{D}}) \, p_{G(\bm{x}_{\mathcal{D}}) | \sigma^2}(\bm{z}) \, p_{\sigma^2}(\sigma^2) \propto \\
		& \propto \frac{1}{\left(\sigma^2\right)^{n_L k / 2}} \exp{-\frac{1}{2\sigma^2} (\bm{y}_{\mathrm{f}} - \bm{z}_{\mathrm{f}})^T (\bm{y}_{\mathrm{f}} - \bm{z}_{\mathrm{f}})} \cdot \\
		& \blankeq \cdot \frac{1}{\left(\sigma^2\right)^{n_L k / 2} \sqrt{\det{\bm{K}'(\bm{x}_{\mathcal{D}})}}} \exp{-\frac{1}{2 \sigma^2} \bm{z}_{\mathrm{f}}^T \left(\bm{K}'(\bm{x}_{\mathcal{D}}) \otimes_K \bm{I}_{n_L}\right)^{-1} \bm{z}_{\mathrm{f}}} \cdot \\
		& \blankeq \cdot \frac{1}{\left(\sigma^2\right)^{a + 1}} \exp{-\frac{b}{\sigma^2}}.
	\end{aligned}
\end{equation*}
Defining $\bm{N} \coloneqq \bm{K}'(\bm{x}_{\mathcal{D}}) \otimes_K \bm{I}_{n_1} \in \operatorname{Sym}_+^{n_1 k}$, $\bm{M} \coloneqq \bm{I}_{n_1 k} + \bm{N}^{-1} \in \operatorname{Sym}_+^{n_1 k}$, through simple manipulations of the exponent we get
\begin{equation*}
	\begin{aligned}
		& (\bm{y}_{\mathrm{f}} - \bm{z}_{\mathrm{f}})^T (\bm{y}_{\mathrm{f}} - \bm{z}_{\mathrm{f}}) + \bm{z}_{\mathrm{f}}^T \bm{N}^{-1} \bm{z}_{\mathrm{f}} = 
		\norm{\bm{y}_{\mathrm{f}}}_2^2 + \norm{\bm{z}_{\mathrm{f}}}_2^2 - 2 \bm{y}_{\mathrm{f}}^T \bm{z}_{\mathrm{f}} + \bm{z}_{\mathrm{f}}^T \bm{N}^{-1} \bm{z}_{\mathrm{f}} = \\
		& \kern20pt = \bm{z}_{\mathrm{f}}^T \left(\bm{I}_{n_L k} + \bm{N}^{-1}\right) \bm{z}_{\mathrm{f}} - 2 \bm{y}_{\mathrm{f}}^T \bm{z}_{\mathrm{f}} + \bm{y}_{\mathrm{f}}^T \bm{y}_{\mathrm{f}} \pm \bm{y}_{\mathrm{f}}^T \left(\bm{I}_{n_L k} + \bm{N}^{-1}\right)^{-1} \bm{y}_{\mathrm{f}} = \\
		& \kern20pt = \bm{z}_{\mathrm{f}}^T \bm{M} \bm{z}_{\mathrm{f}} - 2 \bm{y}_{\mathrm{f}}^T \bm{z}_{\mathrm{f}} + \bm{y}_{\mathrm{f}}^T \bm{y}_{\mathrm{f}} \pm \bm{y}_{\mathrm{f}}^T \bm{M}^{-1} \bm{y}_{\mathrm{f}} = \\
		& \kern20pt = \left(\bm{z}_{\mathrm{f}} - \bm{M}^{-1} \bm{y}_{\mathrm{f}}\right)^T \bm{M} \left(\bm{z}_{\mathrm{f}} - \bm{M}^{-1} \bm{y}_{\mathrm{f}}\right) + \bm{y}_{\mathrm{f}}^T \left(\bm{I}_{n_L k} - \bm{M}^{-1}\right) \bm{y}_{\mathrm{f}}.
	\end{aligned}
\end{equation*}
Substituting and multiplying for the constant term $\sqrt{\det{\bm{M}^{-1}}}$ we obtain
\begin{equation} \label{eq:marginalposteriors}
	\begin{aligned}
		p_{G(\bm{x}_{\mathcal{D}}), \sigma^2 | \train}(\bm{z}, \sigma^2) & \propto \frac{1}{\left(\sigma^2\right)^{n_L k / 2} \sqrt{\det{\bm{M}^{-1}}}} \exp{-\frac{1}{2 \sigma^2} (\bm{z}_{\mathrm{f}} - \bm{M}^{-1} \bm{y}_{\mathrm{f}})^T \bm{M} (\bm{z}_{\mathrm{f}} - \bm{M}^{-1} \bm{y}_{\mathrm{f}})} \cdot \\
		& \blankeq \cdot \frac{1}{\left(\sigma^2\right)^{(a + n_L k / 2) + 1}} \exp{-\frac{1}{\sigma^2} \left(b + \frac{1}{2} \left(\bm{y}_{\mathrm{f}}^T \left(\bm{I}_{n_L k} - \bm{M}^{-1}\right) \bm{y}_{\mathrm{f}}\right)\right)}.
	\end{aligned}
\end{equation}
From \cref{eq:marginalposteriors} it is possible to identify two kernels: one associable with a Gaussian density and the other with an Inverse-Gamma density,
\begin{equation*}
	\begin{aligned}
		\flatten{G(\bm{x}_{\mathcal{D}})} \given \sigma^2, \train & \sim \normal{\bm{M}^{-1} \flatten{\bm{y}_{\mathcal{D}}}}{\sigma^2 \bm{M}^{-1}}, \\
		\sigma^{2} \given \train & \sim \invgamma{a + \frac{n_L k}{2}}{b + \frac{1}{2} \left(\flatten{\bm{y}_{\mathcal{D}}}^T \left(\bm{I}_{n_1 k} - \bm{M}^{-1}\right) \flatten{\bm{y}_{\mathcal{D}}}\right)}.
	\end{aligned}
\end{equation*}
This result allows us to apply the following \Cref{lem:norminvgammastudent} (see \citet{bayesiantheory2009}) and state that the induced posterior distribution, $G(\bm{x}_{\mathcal{D}}) \given \train$ is a $(n_L \times k)$-dimensional Student-$t$:
\begin{equation*}
	\begin{aligned}
		\flatten{G(\bm{x}_{\mathcal{D}})} \given \train \sim \tstud{2a + n_L k}{\bm{\mu}_{\mathrm{post}}}{\bm{\Sigma}_{\mathrm{post}}},
	\end{aligned}
\end{equation*}
with
\begin{equation*}
	\begin{aligned}
		\bm{M} & \coloneqq \bm{I}_{n_L k} + \left(\bm{K}'(\bm{x}) \otimes_K \bm{I}_{n_L}\right)^{-1}, \\
		\bm{\mu}_{\mathrm{post}} & \coloneqq \bm{M}^{-1} \flatten{\bm{y}_{\mathcal{D}}}, \\
		\bm{\Sigma}_{\mathrm{post}} & \coloneqq \left(b + \frac{1}{2} \left(\flatten{\bm{y}_{\mathcal{D}}}^T \left(\bm{I}_{n_L k} - \bm{M}^{-1}\right) \flatten{\bm{y}_{\mathcal{D}}}\right)\right) \frac{2}{2a + n_L k} \bm{M}^{-1}.
	\end{aligned}
\end{equation*}

\begin{lemma} \label{lem:norminvgammastudent}
	Let $k \in \natural_{> 0}$, and $(\bm{z}, \sigma^2)$ Gaussian-Inverse-Gamma (Gaussian-IG) distributed, i.e.,
	\begin{align*}
		\bm{z} \given \sigma^2 & \sim \normal{\bm{\mu}}{\sigma^2 \bm{\Lambda}} \text{ with } \bm{\mu} \in \real^k, \, \bm{\Lambda} \in \operatorname{Sym}_+^{k} \text{ and} \\
		\sigma^2 & \sim \invgamma{\alpha}{\beta} \text{ with } \alpha, \beta > 0,
	\end{align*}
	then $\bm{z}$ is distributed as a $k$-dimensional Student-$t$ with $2 \alpha$ degrees of freedom, $\bm{z} \sim \tstud{2\alpha}{\bm{\mu}}{\frac{\beta}{\alpha} \bm{\Lambda}}$.
\end{lemma}
\begin{proof}
	We know, by hypothesis that $(\bm{z}, \sigma^2)$ is such that
	\begin{align*}
		p_{(\bm{z}, \sigma^2)}(\bm{z}, \sigma^2) & = \frac{1}{(2 \pi)^{\frac{k}{2}} \left(\sigma^2\right)^{\frac{k}{2}} \sqrt{\det{\bm{\Lambda}}}} \exp{-\frac{1}{2 \sigma^2} (\bm{z} - \bm{\mu})^T \bm{\Lambda}^{-1} (\bm{z} - \bm{\mu})} \cdot \\
		& \blankeq \cdot \frac{\beta^\alpha}{\Gamma(\alpha)} \left(\frac{1}{\sigma^2}\right)^{\alpha + 1} \exp{-\frac{\beta}{\sigma^2}}.
	\end{align*}
	Marginalizing over the variance $\sigma^2$ we get
	\begin{equation} \label{eq:ninvgstud1}
		\begin{aligned}
			p_{\bm{z}}(\bm{z}) & = \int_{0}^{\infty} p_{(\bm{z}, \sigma^2)}(\bm{z}, \sigma^2) d \sigma^2 = \\
			& \propto \int_{0}^{\infty} \exp{-\frac{1}{2 \sigma^2} \left(2 \beta + (\bm{z} - \bm{\mu})^T \bm{\Lambda}^{-1} (\bm{z} - \bm{\mu})\right)} \left(\frac{1}{\sigma^2}\right)^{\alpha + \frac{k}{2} + 1} d \sigma^2.
		\end{aligned}
	\end{equation}
	Setting $a = \alpha + \frac{k}{2}$, $b = \frac{2 \beta + (\bm{z} - \bm{\mu})^T \bm{\Lambda}^{-1} (\bm{z} - \bm{\mu})}{2}$, $s = \sigma^2$ one can rewrite the last line of \cref{eq:ninvgstud1} as 
	\begin{equation} \label{eq:ninvgstud2}
		\begin{aligned}
			p_{\bm{z}}(\bm{z}) & \propto \int_{0}^{\infty} s^{-(a + 1)} \exp{-\frac{b}{s}} d s = \int_{\infty}^{0} \left(\frac{t}{b}\right)^{a + 1} e^{-t} \left(-\frac{b}{t^2}\right) dt = \\
			& = \int_{0}^{\infty} b^{-a} t^{a - 1} e^{-t} dt = \Gamma(a) b^{-a} \propto \left(\frac{2 \beta + (\bm{z} - \bm{\mu})^T \bm{\Lambda}^{-1} (\bm{z} - \bm{\mu})}{2}\right)^{-\left(\alpha + \frac{k}{2}\right)} \propto \\
			& \propto \left(1 + \frac{1}{2 \alpha} (\bm{z} - \bm{\mu})^T \left(\frac{\beta}{\alpha} \bm{\Lambda}\right)^{-1} (\bm{z} - \bm{\mu})\right)^{-\frac{2\alpha + k}{2}},
		\end{aligned}
	\end{equation}
	where in the first equality of \cref{eq:ninvgstud2} we performed the change of variable $t = \frac{b}{s}$. In the final form of $p_{\bm{z}}(\bm{z})$ it is possible to recognize the kernel of a $k$-dimensional Student-$t$, $\bm{z} \sim \tstud{2 \alpha}{\bm{\mu}}{\frac{\beta}{\alpha} \bm{\Lambda}}$.
\end{proof}

\section{Proof of the main result} \label{sec:mainproof}

\subsection{Distance between marginal posterior of BNNs and NNGP} \label{subsec:convposteriorsigma}

\begin{proposition} \label{prop:likelihood}
	Let $\mathcal{L}$ be a Gaussian likelihood, $\mathcal{L} \sim \normal{\bm{z}}{\sigmay^2 \bm{I}_{n_L \times k}}$, it holds
	\begin{equation*}
		\norm{\mathcal{L}}_{\infty} = \frac{1}{\left(2 \pi \sigmay^2\right)^{n_L k / 2}} \quad \text{and} \quad \lipschitz{\mathcal{L}} = \frac{e^{-1/2}}{\sqrt{\sigmay^2}} \frac{1}{\left(2 \pi \sigmay^2\right)^{n_L k / 2}}.
	\end{equation*}
\end{proposition}
\begin{proof}
	We first rewrite the map in a more compact form in terms of the Frobenius norm of $(\bm{y}_{\mathcal{D}} - \bm{z})$: $\forall \bm{z} \in \real^{n_L \times k}$,
	\begin{equation*}
		\mathcal{L}(\bm{z}; \bm{y}_{\mathcal{D}}) = \frac{1}{\left(2 \pi \sigmay^2\right)^{n_L k / 2}} \exp{-\frac{1}{2\sigmay^2} \norm{\bm{y}_{\mathcal{D}} - \bm{z}}_F^2}.
	\end{equation*}
	Then for the uniform norm it is sufficient to recall that $\mathcal{L}$ is a bell-shaped map with maximum in the mean point. 
	Therefore, it is immediate that
	\begin{equation*}
		\norm{\mathcal{L}}_{\infty} = \mathcal{L}(\bm{y}_{\mathcal{D}}; \bm{y}_{\mathcal{D}}) = \frac{1}{\left(2 \pi \sigmay^2\right)^{n_L k / 2}}.
	\end{equation*}
	For the identification of the Lipschitz constant it is necessary to recall that, as a consequence of the Mean Value Theorem, given a map $g: \Omega \to \real$ with $\Omega$ open convex subset of $\real^S$, $S$ finite set, if $\sup_{\bm{z} \in \Omega} \norm{\partial / \partial \bm{z} \, g(\bm{z})} \leq L$, then $\mathcal{L}$ is $L$-Lipschitz.
	Hence, our objective is to identify the value of $\sup_{\bm{z} \in \real^{n_L \times k}} \norm{\partial / \partial \bm{z} \, \mathcal{L}(\bm{z}; \bm{y}_{\mathcal{D}})}_F$. \\
	Let us define $c = \frac{1}{\left(2 \pi \sigmay^2\right)^{n_L k / 2}}$, then
	\begin{equation*}
		\begin{aligned}
			\frac{\partial}{\partial \bm{z}} \mathcal{L}(\bm{z}; \bm{y}_{\mathcal{D}}) & = c \, \frac{\partial}{\partial \bm{z}} \exp{-\frac{1}{2\sigmay^2} \norm{\bm{y}_{\mathcal{D}} - \bm{z}}_F^2} = \\
			& = c \, \frac{\partial}{\partial \norm{\bm{y}_{\mathcal{D}} - \bm{z}}_F^2} \exp{-\frac{1}{2\sigmay^2} \norm{\bm{y}_{\mathcal{D}} - \bm{z}}_F^2} \frac{\partial}{\partial (\bm{y}_{\mathcal{D}} - \bm{z})} \norm{\bm{y}_{\mathcal{D}} - \bm{z}}_F^2 \frac{\partial}{\partial \bm{z}} (\bm{y}_{\mathcal{D}} - \bm{z}) = \\
			& = c \, \left(- \frac{1}{2 \sigmay^2} \exp{-\frac{1}{2\sigmay^2} \norm{\bm{y}_{\mathcal{D}} - \bm{z}}_F^2}\right) 2 (\bm{y}_{\mathcal{D}} - \bm{z}) \left(- \bm{I}_{n_L \times k}\right) = \frac{\bm{y}_{\mathcal{D}} - \bm{z}}{\sigmay^2} \mathcal{L}(\bm{z}; \bm{y}_{\mathcal{D}}).
		\end{aligned}
	\end{equation*}
	To find the supremum of $h: \real^{n_L \times k} \to \real$, 
	\begin{equation*}
		h(\bm{z}) \coloneqq \norm{\frac{\partial}{\partial \bm{z}} \mathcal{L}(\bm{z}; \bm{y}_{\mathcal{D}})}_F = \frac{\norm{\bm{y}_{\mathcal{D}} - \bm{z}}_F}{\sigmay^2} \mathcal{L}(\bm{z}; \bm{y}_{\mathcal{D}}), \ \forall \bm{z} \in \real^{n_L \times k},
	\end{equation*}
	we first have to notice that $h$ is a positive real valued map, and that its first derivative has zeros in every $\bm{z}_0$ such that $\norm{\bm{y}_{\mathcal{D}} - \bm{z}_0}_F^2 = \sigmay^2$, which, as a consequence, are critical points. Indeed, for every $\bm{z} \neq \bm{y}_{\mathcal{D}}$,
	\begin{align*}
		\frac{\partial}{\partial \bm{z}} h(\bm{z}) & = \frac{1}{\sigmay^2} \left(\frac{\partial}{\partial \bm{z}} \mathcal{L}(\bm{z}; \bm{y}_{\mathcal{D}}) \norm{\bm{y}_{\mathcal{D}} - \bm{z}}_F + \mathcal{L}(\bm{z}; \bm{y}_{\mathcal{D}}) \frac{\partial}{\partial \bm{z}} \norm{\bm{y}_{\mathcal{D}} - \bm{z}}_F\right) = \\
		& = \frac{1}{\sigmay^2} \left(\frac{\bm{y}_{\mathcal{D}} - \bm{z}}{\sigmay^2} \mathcal{L}(\bm{z}; \bm{y}_{\mathcal{D}}) \norm{\bm{y}_{\mathcal{D}} - \bm{z}}_F - \mathcal{L}(\bm{z}; \bm{y}_{\mathcal{D}}) \frac{\bm{y}_{\mathcal{D}} - \bm{z}}{\norm{\bm{y}_{\mathcal{D}} - \bm{z}}_F}\right) = \\
		& = \frac{1}{\sigmay^2} \mathcal{L}(\bm{z}; \bm{y}_{\mathcal{D}}) \frac{\bm{y}_{\mathcal{D}} - \bm{z}}{\norm{\bm{y}_{\mathcal{D}} - \bm{z}}_F} \left(\frac{\norm{\bm{y}_{\mathcal{D}} - \bm{z}}_F^2}{\sigmay^2} - 1\right),
	\end{align*}
	which is null if and only if $\norm{\bm{y}_{\mathcal{D}} - \bm{z}}_F^2 / \sigmay^2 = 1$. \\
	For any such $\bm{z}_0$ we would have that
	\begin{equation*}
		h(\bm{z}_0) = \mathcal{L}\left(\bm{z}_0; \bm{y}_{\mathcal{D}}\right) = \frac{1}{\sqrt{\sigmay^2}} \frac{e^{-1/2}}{\left(2 \pi \sigmay^2\right)^{n_L k / 2}}.
	\end{equation*}
	It is also possible to define 
	\begin{equation*}
		l: \real^+ \to \real, \ l(x) \coloneqq \frac{x}{\sigmay^2} \frac{1}{\left(2 \pi \sigmay^2\right)^{n_L k / 2}} \exp{-\frac{1}{2 \sigmay^2} x^2}, \ \forall x \in \real^+,
	\end{equation*}
	which, by construction, has the following property: 
	\begin{equation*}
		\forall \bm{z} \in \real^{n_L \times k}, \ l(\norm{\bm{y}_{\mathcal{D}} - \bm{z}}_F) = h(\bm{z}).
	\end{equation*}
	From the definition it is easy to see that $\sup_{x \in \real^+} l(x) = h(\bm{z}_0)$, just computing its first two derivatives $l'$ and $l''$. Defining $d = \sigmay^{-2} \left(2 \pi \sigmay^2\right)^{- n_L k / 2}$ positive constant we have
	\begin{align*}
		l'(x) & = d \exp{-\frac{1}{2\sigmay^2} x^2} \left(1 - x^2 \frac{1}{\sigmay^2}\right), \\
		l''(x) & = - d \exp{-\frac{1}{2\sigmay^2} x^2} \frac{1}{\sigmay^2} x \left(\left(1 - x^2 \frac{1}{\sigmay^2}\right) + 2\right) = \\
		& = - d \exp{-\frac{1}{2\sigmay^2} x^2} \frac{1}{\sigmay^2} x \left(3 - \frac{x^2}{\sigmay^2}\right),
	\end{align*}
	and so
	\begin{align*}
		l'(x) = 0 \iff x^2 = \sigmay^2, \ l''(x) \leq 0 \iff x^2 \leq 3 \sigmay^2.
	\end{align*}
	This implies that in $x = \sqrt{\sigmay^2}$ there is a local maximum and observing the asymptotic behavior, $\lim_{x \to +\infty} l(x) = 0$, we know that it is the unique global maximum. 
	Finally, evaluating $l$ in $\sqrt{\sigmay^2}$ we get 
	\begin{equation*}
		l\left(\sqrt{\sigmay^2}\right) = \frac{1}{\sqrt{\sigmay^2}} \frac{e^{- 1/2}}{\left(2 \pi \sigmay^2\right)^{n_L k / 2}}.
	\end{equation*}
	Therefore, it follows that the critical points $\bm{z}_0$ are global maxima for $h$ and so
	\begin{equation*}
		\sup_{\bm{z} \in \real^{n_L \times k}} \norm{\frac{\partial}{\partial \bm{z}} \mathcal{L}(\bm{z}; \bm{y}_{\mathcal{D}})}_F = h(\bm{z}_0).
	\end{equation*}
\end{proof}

For the sake of completeness we report below a more detailed version of the Corollary 5.3 of \citet{trevisan2023}, which we use as a starting point for the subsequent results.

\begin{corollary} \label{cor:convposterior}
	Given $f_{\bm{\theta}}$ BNN, with architecture $\bm{\alpha} = (\bm{n}, \bm{\varphi})$, $\bm{\varphi}$ collection of Lipschitz activation functions, prior distribution on $\bm{\theta}$ as in \cref{eq:nealprior}, $G$ Gaussian process as in \cref{eq:nngp}, $\bm{x}$, $\bm{y}_{\mathcal{D}}$ as in \Cref{subsec:postbnn}, and a Gaussian likelihood function $\mathcal{L} \sim \normal{\bm{z}}{\sigmay^2 \bm{I}_{n_L \times k}}$, exists a constant
	\begin{equation*}
		c\left(\train, \bm{\varphi}, \bm{\sigma}, \sigmay^2, n_L\right) > 0, \text{ independent of } \left(n_l\right)_{l = 1}^{L - 1},
	\end{equation*}
	such that,
	\begin{equation*}
		\wass[1]{f_{\bm{\theta}}(\bm{x}) \given \train}{G(\bm{x}) \given \train} \leq c \frac{1}{\sqrt{n_{min}}},
	\end{equation*}
	for all $\displaystyle n_{min} \coloneqq \min_{l = 1, \dots, L - 1} n_l$ sufficiently large.
\end{corollary}
\begin{proof}
	Let $\widetilde{\mu}$ the law of the induced prior distribution of a BNN, $\widetilde{\mu} \sim f_{\bm{\theta}}(\bm{x})$, and $\mu$ be the law of the associated NNGP, $\mu \sim G(\bm{x})$, probability measures on $\left(\real^{n_L \times k}, \norm{\cdot}\right)$. \\
	Let $g \coloneqq \mathcal{L}: \real^{n_L \times k} \to \real$, bounded Lipschitz map.
	The posterior distributions $\mathbb{P}_{f_{\bm{\theta}}(\bm{x}) | \train}$ and $\mathbb{P}_{G(\bm{x}) | \train}$ are, by construction, respectively equal to $\widetilde{\mu}_g$ and $\mu_g$ as they are defined in \Cref{lem:nnposterior}, therefore, by a direct application with $p = 2$, we get the following rewriting of \cref{eq:wassposterior}:
	\begin{equation} \label{eq:wasspostcor}
		\begin{aligned}
			& \wass[1]{f_{\bm{\theta}}(\bm{x}) \given \train}{G(\bm{x}) \given \train} \leq \frac{1}{p^{(1)}} \left(\lipschitz{\mathcal{L}} p^{(3)} + \left(1 + \frac{p^{(4)} \lipschitz{\mathcal{L}}}{p^{(2)}}\right) \norm{\mathcal{L}}_{\infty}\right) \wass[2]{f_{\bm{\theta}}(\bm{x})}{G(\bm{x})},
		\end{aligned}
	\end{equation}
	where $\lipschitz{\mathcal{L}}, \norm{\mathcal{L}}_{\infty}$ are positive constants depending on $k, \sigmay^2, n_L$, as showed in \Cref{prop:likelihood}, and
	\begin{equation*}
		\begin{gathered}
			p^{(1)} = \mean[\bm{z} \sim G(\bm{x})]{\mathcal{L}(\bm{z}; \bm{y}_{\mathcal{D}})}, \ p^{(2)} = \mean[\bm{z} \sim f_{\bm{\theta}}(\bm{x})]{\mathcal{L}(\bm{z}; \bm{y}_{\mathcal{D}})}, \\
			p^{(3)} = \mean[\bm{z} \sim G(\bm{x})]{\norm{\bm{z}}_F^{2}}, \ p^{(4)} = \mean[\bm{z} \sim G(\bm{x})]{\norm{\bm{z}}_F}.
		\end{gathered}
	\end{equation*}
	It is immediate that $p^{(3)}$ and $p^{(4)}$ are finite indeed they are the second and the first moment of a multivariate centered Gaussian, $G(\bm{x}) \sim \normal{\bm{0}_{n_L \times k}}{\bm{I}_{n_L} \otimes \bm{K}(\bm{x})}$, respectively. Therefore, as we observed in \Cref{rem:gaussmoments}, it holds
	\begin{align} \label{eq:p3-4}
	(p^{(4)})^2 \leq p^{(3)} & = \trace{\bm{I}_{n_L} \otimes \bm{K}(\bm{x})} = n_L \trace{\bm{K}(\bm{x})} \leq n_L k \max_{i \in [k]} \bm{K}(\bm{x}_i, \bm{x}_i) \leq \nonumber \\
		& \leq n_L k \left(\sigma_{\bm{W}^{(L)}}^2 \max_{i \in [k]} \mean{\norm{\varphi_{L}\left(G^{(L - 1)}(\bm{x}_i)\right)}_2^2} / n_L + \sigma_{\bm{b}^{(L)}}^2\right).
	\end{align} 
	So both the terms can be bounded by expressions only dependent on $\bm{x}, \bm{\varphi}, \bm{\sigma}$ and $n_L$. \\
	For the remaining terms we still exploit the normality of $G(\bm{x})$ (see \cref{eq:nngpx}). 
	Due to the boundedness and positivity of $\mathcal{L}: \real^{d_{\mathrm{out} \times k}} \to (0, \norm{\mathcal{L}}_{\infty})$, it is clear that $p^{(1)} \in (0, \infty)$, indeed we are integrating $\mathcal{L}$ with respect to a strictly positive probability measure: $0 < p^{(1)}(\train, \bm{\varphi}, \bm{\sigma}, \sigmay^2, n_L) =$ $= \mean{\mathcal{L}(G(\bm{x}); \bm{y}_{\mathcal{D}})} \leq \norm{\mathcal{L}}_{\infty}$. \\
	Moreover, it is possible to notice that considering $\bm{v} \sim G(\bm{x})$, $\bm{w} \sim f_{\bm{\theta}}(\bm{x})$,
	\begin{align*}
		|p^{(1)} - p^{(2)}| & = \left|\mean{\mathcal{L}(\bm{v}; \bm{y}_{\mathcal{D}}) - \mathcal{L}(\bm{w}; \bm{y}_{\mathcal{D}})}\right| \leq \norm{\mathcal{L}(\bm{v}; \bm{y}_{\mathcal{D}}) - \mathcal{L}(\bm{w}; \bm{y}_{\mathcal{D}})}_{L^1} \leq \lipschitz{\mathcal{L}} \norm{\bm{v} - \bm{w}}_{L^1},
	\end{align*}
	and taking the infimum over the couplings $(\bm{v}, \bm{w})$ we get
	\begin{align} \label{eq:p1-2}
		|p^{(1)} - p^{(2)}| \leq \lipschitz{\mathcal{L}} \wass[1]{G(\bm{x})}{f_{\bm{\theta}}(\bm{x})} \leq \lipschitz{\mathcal{L}} c_1 \frac{1}{\sqrt{n_{min}}},
	\end{align}
	where for the last inequality we used a compact version of the result in \Cref{thm:priornngp}. Recalling that $c_1$ and $\lipschitz{\mathcal{L}}$ depend only on $\bm{x}, \bm{\varphi}, \bm{\sigma}, \sigmay^2$ and $n_L$, \cref{eq:p1-2} implies that, for $n_{min}$ sufficiently large, $p^{(2)}$  is also strictly positive. \\
	To conclude it is sufficient to note that \cref{eq:wasspostcor}, together with \Cref{thm:priornngp} and the observations made about $(p^{(i)})_{i = 1}^4$ lead to
	\begin{align*}
		\wass[1]{f_{\bm{\theta}}(\bm{x}) \given \train}{G(\bm{x}) \given \train} & \leq c_2 \wass[2]{f_{\bm{\theta}}(\bm{x})}{G(\bm{x})} \leq c_2 c_3 \sqrt{n_L} \sum_{l = 1}^{L} \frac{1}{\sqrt{n_k}} \leq c \frac{1}{\sqrt{n_{min}}},
	\end{align*}
	where $c_2$ and $c_3$ depends on $\train, \bm{\varphi}, \bm{\sigma}, \sigmay^2, n_L$.
\end{proof}

\begin{lemma} \label{lem:priornngpsigma}
	Given $f_{\bm{\theta}}$, $G$ and $\bm{x}$ as in \Cref{thm:priornngp}, $p \geq 1$, and assuming $\sigma^2 \coloneqq \sigma_{\bm{W}^{(L)}}^2 =$ $= \sigma_{\bm{b}^{(L)}}^2 = \sigmay^2$, there exists a constant
	\begin{equation*}
		c\left(p, \bm{x}, \bm{\varphi}, (\bm{\sigma}_l)_{l = 1}^{L - 1}\right) > 0, \text{ independent of } \left(n_l\right)_{l = 1}^L, \sigma^2,
	\end{equation*}
	such that
	\begin{equation*}
		\wass[p]{f_{\bm{\theta}}(\bm{x}) \given \sigma^2}{G(\bm{x}) \given \sigma^2} \leq \sigma c \sqrt{n_L} \sum_{l = 1}^{L - 1} \frac{1}{\sqrt{n_l}},
	\end{equation*}
\end{lemma}
\begin{proof}
	The proof is straightforward if we observe that, by construction (see \cref{eq:nn,eq:nngpx}) $f_{\bm{\theta}} = \sigma f_{\bm{\theta}'}$ and $G = \sigma  G'$, with 
	\begin{equation} \label{eq:thetaprime}
		\bm{\theta}' \text{ such that } \bm{\sigma}' = \left(\left(\left(\sigma_{\bm{W}^{(l)}}^2, \sigma_{\bm{b}^{(l)}}^2\right)\right)_{l = 1}^{L - 1}, (1, 1)\right),
	\end{equation}
	and $G'$ built with weights and bias variances as in $\bm{\sigma}'$. \\
	Indeed, applying \cref{eq:wasslinear} and \Cref{thm:priornngp} we get
	\begin{equation*}
		\wass[p]{f_{\bm{\theta}}(\bm{x}) \given \sigma^2}{G(\bm{x}) \given \sigma^2} =\sigma \wass[p]{f_{\bm{\theta}}'(\bm{x})}{G'(\bm{x})} \leq \sigma c \sqrt{n_L} \sum_{l = 1}^{L - 1} \frac{1}{\sqrt{n_L}}, 
	\end{equation*}
	with $c$ independent of $\sigma^2$.
\end{proof}

\begin{lemma} \label{lem:boundexplikelihood}
	Given $f_{\bm{\theta}}$, $G$, $\bm{x}$, $\bm{y}_{\mathcal{D}}$ and $\mathcal{L}$ as in \Cref{cor:convposterior}, assuming $\sigma^2 \coloneqq \sigma_{\bm{W}^{(L)}}^2 = \sigma_{\bm{b}^{(L)}}^2 =$ $= \sigmay^2$, $\bm{K}'(\bm{x}) \in \operatorname{Sym}_+^{k}$ (rescaled NNGP kernel), and 
	\begin{equation*}
		\bm{v} \sim G(\bm{x}) \given \sigma^2, \, \bm{w} \sim f_{\bm{\theta}}(\bm{x}) \given \sigma^2,
	\end{equation*}
	$\forall \epsilon < 1/\norm{\bm{K}'(\bm{x})}_{\mathrm{op}}$ it holds
	\begin{equation*}
		\begin{aligned}
			c_1 \cdot \left(\sigma^2\right)^{- n_L k / 2} \exp{-\frac{\norm{\bm{y}_{\mathcal{D}}}_F^2}{\sigma^2}} \leq \ & \mean{\mathcal{L}(\bm{v}; \bm{y}_{\mathcal{D}})} \leq c_2 \cdot \left(\sigma^2\right)^{- n_L k / 2} \exp{-\frac{1}{2 \sigma^2} \frac{\epsilon}{\epsilon + 1} \norm{\bm{y}_{\mathcal{D}}}_F^2}, \\
			c_3 \exp{-\frac{\norm{\bm{y}_{\mathcal{D}}}_F^2}{\sigma^2}} \leq \ & \mean{\mathcal{L}(\bm{w}; \bm{y}_{\mathcal{D}})} \leq c_4 \cdot \left(\sigma^2\right)^{- n_L k / 2},
		\end{aligned}
	\end{equation*}
	where $n_{min} = \min_{l = 1, \dots, L - 1} n_l$ is sufficiently large and the constants depend on $\bm{x}, \bm{\varphi}$, $\left(\bm{\sigma}_l\right)_{l = 1}^{L - 1}, n_L$.
\end{lemma}
\begin{proof}
	We separately discuss the four bounds.
	
	\textbf{Bounds related to $G(\bm{x})$.}	It is possible to rewrite $\mean{\mathcal{L}(\bm{v}; \bm{y}_{\mathcal{D}})}$ as follows:
	\begin{align} \label{eq:nngpexplikelihood}
		\mean{\mathcal{L}(\bm{v}; \bm{y}_{\mathcal{D}})} & \propto \mean{\left(\sigma^2\right)^{- n_L k / 2} \exp{-\frac{1}{2\sigma^2} \norm{\bm{y}_{\mathcal{D}} - \bm{v}}_F^2}} = \nonumber \\
		& = \left(\sigma^2\right)^{- n_L k / 2} \int_{\real^{n_L \times k}} \exp{-\frac{1}{2\sigma^2} \norm{\bm{y}_{\mathcal{D}} - \bm{v}}_F^2} d \mathbb{P}_{G(\bm{x}) | \sigma^2}(\bm{v}).
	\end{align}
	Starting with the density of the Gaussian variable $G(\bm{x})$,
	\begin{equation*}
		\begin{aligned}	
			p_{G(\bm{x})}(\bm{z}) & = \frac{1}{\left((2\pi)^{n_L k} \det{\bm{K}(\bm{x}) \otimes_K \bm{I}_{n_L}}\right)^{1/2}} \cdot \\
			& \blankeq \cdot \exp{-\frac{1}{2} \flatten{\bm{z}}^T \left(\bm{K}(\bm{x}) \otimes_K \bm{I}_{n_L} \right)^{-1} \flatten{\bm{z}}},
		\end{aligned}
	\end{equation*}
	and recalling that, $\forall \bm{A} \in \real^{n \times n}, \bm{B} \in \real^{k \times k}$, then $(\bm{A} \otimes_K \bm{B})^{-1} = \bm{A}^{-1} \otimes_K \bm{B}^{-1}$, $\det{\bm{A} \otimes_K \bm{B}} =$ $= \det{\bm{A}}^k \det{\bm{B}}^n$, and $\bm{K}(\bm{x}) = \sigma^2 \bm{K}'(\bm{x})$, with $\bm{K}'(\bm{x})$ as in \cref{eq:rescaledkernel}, it is easy to see that 
	\begin{equation} \label{eq:densitynngp}
		\begin{aligned}	
			p_{G(\bm{x}) | \sigma^2}(\bm{v}) = \frac{1}{(2 \pi \sigma^2)^{n_L k / 2}} \frac{1}{\det{\bm{K}'(\bm{x})}^{n_L / 2}} \exp{-\frac{1}{2 \sigma^2} \sum_{i, j \in [k] \times [k]} \left(\bm{K}'(\bm{x})^{-1}\right)_{i, j} \bm{v}_i^T \bm{v}_j}.
		\end{aligned}
	\end{equation}

	\textbf{Lower bound.} Substituting \cref{eq:densitynngp} in \cref{eq:nngpexplikelihood} we get
	\begin{align} 
		& \begin{aligned} \label{eq:nngpexplikprop}
			\mean{\mathcal{L}(\bm{v}; \bm{y}_{\mathcal{D}})} & \propto \left(\sigma^2\right)^{- n_L k} \int_{\real^{n_L \times k}} \exp{-\frac{1}{2\sigma^2} \norm{\bm{y}_{\mathcal{D}} - \bm{v}}_F^2} \cdot \\
			& \blankeq \cdot \exp{-\frac{1}{2\sigma^2} \sum_{i, j \in [k] \times [k]} \left(\bm{K}'(\bm{x})^{-1}\right)_{i, j} \bm{v}_i^T \bm{v}_j} d\bm{v} \geq 
		\end{aligned} \\
		& \kern49.75pt \begin{aligned} \label{eq:nngpexplikgeq}
			& \geq \left(\sigma^2\right)^{- n_L k} \exp{-\frac{1}{\sigma^2} \norm{\bm{y}_{\mathcal{D}}}_F^2} \left(\sigma^2\right)^{n_L k /2} \cdot \\
			& \blankeq \cdot \int_{\real^{n_L \times k}} \exp{-\frac{1}{2} \left(\sum_{i, j \in [k] \times [k]} \left(\bm{K}'(\bm{x})^{-1} + 2 \cdot \bm{I}_k\right)_{i, j} \bm{u}_i^T \bm{u}_j\right)} d\bm{u},
		\end{aligned}
	\end{align}
	where the inequality follows applying $\norm{\bm{y}_{\mathcal{D}} - \bm{v}}_F^2 \leq 2(\norm{\bm{y}_{\mathcal{D}}}_F^2 + \norm{\bm{v}}_F^2)$ (see \cref{eq:ineqnormsqleq}) and performing the change of variable $\bm{u} \coloneqq \bm{v} / (\sigma^2)^{1/2}$. 
	The conclusion for the lower bound follows easily observing that $\bm{K}'(\bm{x})$ is positive definite, so it holds 
	\begin{equation*}
		(\bm{K}'(\bm{x})^{-1} + 2 \cdot \bm{I}_k) \otimes_K \bm{I}_{n_L} \in \operatorname{Sym}_+^{n_L k}, 
	\end{equation*}
	and therefore given $\bm{u}_{\mathrm{f}} = \flatten{\bm{u}}$ the integral in \cref{eq:nngpexplikgeq} is equal to a constant depending on $\bm{x}, \bm{\varphi}, \left(\bm{\sigma}_l\right)_{l = 1}^{L - 1}, n_L$: 
	\begin{equation} \label{eq:boundinfthetaprime}
		\begin{aligned}
			& \int_{\real^{n_L \times k}} \exp{-\frac{1}{2} (\bm{u}_{\mathrm{f}}^T \left((\bm{K}'(\bm{x})^{-1} + 2 \cdot \bm{I}_k) \otimes_K \bm{I}_{n_L}\right) \bm{u}_{\mathrm{f}})} d\bm{u}_{\mathrm{f}} = \\
			& \kern20pt = \left((2\pi)^{k} \det{\bm{K}'(\bm{x})^{-1} + 2 \cdot \bm{I}_k}\right)^{n_L / 2}.
		\end{aligned}
	\end{equation}   
	
	\textbf{Upper bound.} For the upper bound the procedure is similar.
	Starting from the result in \cref{eq:nngpexplikprop}, applying the inequality $\norm{\bm{y}_{\mathcal{D}} - \bm{v}}_F^2 \geq \frac{\epsilon}{1 + \epsilon} \norm{\bm{y}_{\mathcal{D}}}_F^2 - \epsilon \norm{\bm{v}}_F^2$, for a fixed $\epsilon > 0$ (see \cref{eq:ineqnormsqgeqeps}) and performing again the change of variable $\bm{u} \coloneqq \bm{v} / (\sigma^2)^{1/2}$ we get
	\begin{align*}
		\mean{\mathcal{L}(\bm{v}; \bm{y}_{\mathcal{D}})} & \leq c \cdot \left(\sigma^2\right)^{- n_L k} \exp{-\frac{1}{2 \sigma^2} \frac{\epsilon}{\epsilon + 1} \norm{\bm{y}_{\mathcal{D}}}_F^2} \left(\sigma^2\right)^{n_L k /2} \cdot \\
		& \blankeq \cdot \int_{\real^{n_L \times k}} \exp{-\frac{1}{2} \sum_{i, j \in [k] \times [k]} \left(\bm{K}'(\bm{x})^{-1} - \epsilon \bm{I}_k\right)_{i, j} \bm{v}_i^T \bm{v}_j} d\bm{v}.
	\end{align*}
	Now, in order to have a convergent integral it is sufficient to impose the matrix $\bm{K}'(\bm{x})^{-1} - \epsilon \bm{I}_k$ to be positive definite. Indeed, in that case, we could conclude as before. 
	However, this condition is obviously verified 
	\begin{equation*}
		\forall \epsilon < \min\left\{\lambda \given \lambda \in \operatorname{Sp}\left(\bm{K}'(\bm{x})^{-1}\right)\right\} = \max\left\{\lambda \given \lambda \in \operatorname{Sp}\left(\bm{K}'(\bm{x})\right)\right\}^{-1} = 1/\norm{\bm{K}'(\bm{x})}_{\mathrm{op}},
	\end{equation*}
	indeed for such $\epsilon$ one has that 
	\begin{equation*}
		\operatorname{Sp}\left(\bm{K}'(\bm{x})^{-1} - \epsilon \bm{I}_k\right) = \operatorname{Sp}\left(\bm{K}'(\bm{x})^{-1}\right) - \epsilon \subset \real^+.
	\end{equation*}
	
	\textbf{Bounds related to $f_{\bm{\theta}}(\bm{x})$.} As for the Gaussian process we can write 
	\begin{equation} \label{eq:nnexplikelihood}
		\mean{\mathcal{L}(\bm{w}; \bm{y}_{\mathcal{D}})} \propto \left(\sigma^2\right)^{- n_L k / 2} \int_{\real^{n_L \times k}} \exp{-\frac{1}{2\sigma^2} \norm{\bm{y}_{\mathcal{D}} - \bm{w}}_F^2} d \mathbb{P}_{f_{\bm{\theta}}(\bm{x}) | \sigma^2}(\bm{w}).
	\end{equation}
	We also recall the definitions of $f_{\bm{\theta}'} = (\sigma^2)^{-1/2} f_{\bm{\theta}}$ and $G' = (\sigma^2)^{-1/2} G$, respectively the rescaled BNN and NNGP, random processes independent on $\sigma^2$, as in \cref{eq:thetaprime}.

	\textbf{Lower bound.} Applying the same inequality and the same change of variable used in the lower bound related to $G(\bm{x})$ we get 
	\begin{align} \label{eq:nnexplikgeq}
		\mean{\mathcal{L}(\bm{w}; \bm{y}_{\mathcal{D}})} & \geq c \exp{-\frac{1}{\sigma^2} \norm{\bm{y}_{\mathcal{D}}}_F^2} \int_{\real^{n_L \times k}} \exp{-\norm{\bm{u}}_F^2} d \mathbb{P}_{f_{\bm{\theta}}'(\bm{x})}(\bm{u}) \nonumber = \\
		& = c \exp{-\frac{1}{\sigma^2} \norm{\bm{y}_{\mathcal{D}}}_F^2} \mean[\bm{u} \sim f_{\bm{\theta}}'(\bm{x})]{e^{-\norm{\bm{u}}_F^2}}.
	\end{align}
	Now, we can exploit the fact that we know how to integrate $e^{-\norm{\bm{\cdot}}_F^2}$ with respect to the measure $\mathbb{P}_{G'(\bm{x})}$ (we already computed this integral up to a constant depending on the usual parameters, in \cref{eq:boundinfthetaprime}) to obtain analogous results for the mean value in \cref{eq:nnexplikgeq}. 
	As in \cref{eq:p1-2}, exploiting \Cref{thm:priornngp}, and observing that
	\begin{align*}
		\lipschitz{e^{-\norm{\bm{\cdot}}_F^2}} & = \max_{\bm{u} \in \real^{n_L \times k}} \norm{\frac{\partial}{\partial \bm{u}} e^{-\norm{\bm{u}}_F^2}}_F = \max_{\bm{u} \in \real^{n_L \times k}} \norm{-2 \bm{u} e^{-\norm{\bm{u}}_F^2}}_F = \sqrt{\frac{2}{e}},
	\end{align*}
	it is possible to write the following upper bound to the difference of the mean values of $e^{-\norm{\bm{\cdot}}_F^2}$ with respect to the laws of $G'(\bm{x})$ and $f_{\bm{\theta}'}(\bm{x})$:
	\begin{equation} \label{eq:boundmeanexpscaled}
		\begin{aligned}
			& \left|\mean[\bm{u} \sim G'(\bm{x})]{e^{-\norm{\bm{u}}_F^2}} - \mean[\bm{u} \sim f_{\bm{\theta}}'(\bm{x})]{e^{-\norm{\bm{u}}_F^2}}\right| \leq \lipschitz{e^{-\norm{\bm{\cdot}}_F^2}} \wass[1]{G'(\bm{x})}{f_{\bm{\theta}'}(\bm{x})} \leq \frac{c}{\sqrt{n_{min}}}.
		\end{aligned}
	\end{equation}
	Therefore, assuming 
	\begin{equation*}
		\sqrt{n_{min}} \geq 2c / \mean[\bm{u} \sim G'(\bm{x})]{e^{-\norm{\bm{u}}_F^2}},
	\end{equation*}
	we get 
	\begin{equation*}
		\mean[\bm{u} \sim f_{\bm{\theta}}'(\bm{x})]{e^{-\norm{\bm{u}}_F^2}} \geq \mean[\bm{u} \sim G'(\bm{x})]{e^{-\norm{\bm{u}}_F^2}} / 2,
	\end{equation*}
	and therefore
	\begin{equation*}
		\mean{\mathcal{L}(\bm{w}; \bm{y}_{\mathcal{D}})} \geq c \exp{-\frac{1}{\sigma^2} \norm{\bm{y}_{\mathcal{D}}}_F^2}.
	\end{equation*}

	\textbf{Upper bound.} The upper bound can easily be obtained observing that the negative exponential in \cref{eq:nnexplikelihood} is smaller than $1$ and therefore its integral is smaller than $1$ as well\footnote{To find a sharper upper bound reproducing the result just showed for the lower bound is not trivial. We cannot apply an analogue of \cref{eq:boundmeanexpscaled} because the map $e^{\frac{\epsilon}{2} \norm{\bm{\cdot}}_F^2}$ is not Lipschitz.}.
\end{proof}

It is now possible to state and prove the following \Cref{cor:convposteriorsigma}, which is a version of \Cref{cor:convposterior} in which the dependence on $\sigma^2$ is explicit, under the assumptions of BNN built using the hierarchical model defined in \cref{eq:hiermodel}.

\begin{corollary} \label{cor:convposteriorsigma}
	Given $f_{\bm{\theta}}$, $G$, $\bm{x}$, $\bm{y}_{\mathcal{D}}$ and $\mathcal{L}$ as in \Cref{cor:convposterior}, and assuming $\sigma^2 \coloneqq \sigma_{\bm{W}^{(L)}}^2 =$ $= \sigma_{\bm{b}^{(L)}}^2 = \sigmay^2$ exist some constants
	\begin{equation*}
		c_i\left(\train, \bm{\varphi}, (\bm{\sigma}_l)_{l = 1}^{L - 1}, n_L\right) > 0, \,i = 0, \dots, 4, \text{ independent of } \left(n_l\right)_{l = 1}^{L - 1}, \sigma^2,
	\end{equation*}
	such that,
	\begin{equation*}
		\wass[1]{f_{\bm{\theta}}(\bm{x}) \given (\train, \sigma^2)}{G(\bm{x}) \given (\train, \sigma^2)} \leq h(\sigma^2) \frac{c_0}{\sqrt{n_{min}}},
	\end{equation*}
	with
	\begin{equation*}
		h(\sigma^2) = c_1 \left(\sigma^2\right)^{1/2} \exp{\frac{\norm{\bm{y}_{\mathcal{D}}}_F^2}{\sigma^2}} \left(c_2 + c_3 \left(\sigma^2\right)^{1/2} + c_4 \left(\sigma^2\right)^{- n_L k / 2} \exp{\frac{\norm{\bm{y}_{\mathcal{D}}}_F^2}{\sigma^2}}\right).
	\end{equation*}
\end{corollary}
\begin{proof}
	In analogy with the proof of \Cref{cor:convposterior}, the idea is to apply \Cref{lem:nnposterior} with
	\begin{equation*}
		\widetilde{\mu} \sim f_{\bm{\theta}}(\bm{x}) \given \sigma^2, \, \mu \sim G(\bm{x}) \given \sigma^2, \, g \coloneqq \mathcal{L}(\bm{z}; \bm{y}_{\mathcal{D}}, \sigma^2),
	\end{equation*}
	so that
	\begin{equation*}
		\widetilde{\mu}_g = \frac{g}{\widetilde{\mu}(g)} \widetilde{\mu} = \mathbb{P}_{f_{\bm{\theta}}(\bm{x}) \given (\sigma^2, \train)} \quad \text{and} \quad \mu_g = \frac{g}{\mu(g)} \mu = \mathbb{P}_{G(\bm{x}) \given (\sigma^2, \train)}.
	\end{equation*}
	By a direct application of \cref{eq:wassposterior} we get
	\begin{equation} \label{eq:nngpsigma1}
		\begin{aligned}
			& \wass[1]{f_{\bm{\theta}}(\bm{x}) \given (\sigma^2, \train)}{G(\bm{x}) \given (\sigma^2, \train)} \leq \\
			& \kern20pt \leq \frac{1}{p^{(1)}} \left(\lipschitz{\mathcal{L}} p^{(3)} + \left(1 + \frac{p^{(4)} \lipschitz{\mathcal{L}}}{p^{(2)}}\right) \norm{\mathcal{L}}_{\infty}\right) \wass[2]{f_{\bm{\theta}}(\bm{x}) \given \sigma^2}{G(\bm{x}) \given \sigma^2},
		\end{aligned}
	\end{equation}
	where $\norm{\mathcal{L}}_{\infty}$, $\lipschitz{\mathcal{L}}$ are reported explicitly in \Cref{prop:likelihood}, whereas $1 / p^{(1)}$, $1 / p^{(2)}$ and $p^{(3)}$, $p^{(4)}$ are upper bounded respectively in \Cref{lem:boundexplikelihood} and \cref{eq:p3-4}: all the constants depends on $\bm{x}, \bm{\varphi}, (\bm{\sigma}_l)_{l = 1}^{L - 1}$ and $n_L$
	\begin{equation} \label{eq:p1-4final}
		\begin{gathered}
			\norm{\mathcal{L}}_{\infty} = c \, \left(\sigma^2\right)^{- n_L k / 2}, \ \lipschitz{\mathcal{L}} = c \, \left(\sigma^2\right)^{-n_L k / 2 - 1/2}, \\
			p^{(1)} \geq c \, \left(\sigma^2\right)^{- n_L k / 2} \exp{-\frac{\norm{\bm{y}_{\mathcal{D}}}_F^2}{\sigma^2}}, \ p^{(2)} \geq c \, \exp{-\frac{\norm{\bm{y}_{\mathcal{D}}}_F^2}{\sigma^2}}, \\
			p^{(3)} \leq c \, \sigma^2, \ p^{(4)} \leq c \, \left(\sigma^2\right)^{1/2}.
		\end{gathered}
	\end{equation}
	Hence, substituting the results in \cref{eq:p1-4final} inside \cref{eq:nngpsigma1} and applying \Cref{lem:priornngpsigma} we obtain
	\begin{equation*}
		\begin{gathered}
			\wass[1]{f_{\bm{\theta}}(\bm{x}) \given (\sigma^2, \train)}{G(\bm{x}) \given (\sigma^2, \train)} \leq h(\sigma^2) \frac{c}{\sqrt{n_{min}}}, \text{ with} \\
			\begin{aligned}
				h(\sigma^2) & = \left(\sigma^2\right)^{1/2} \left(\sigma^2\right)^{n_L k / 2} \exp{\frac{\norm{\bm{y}_{\mathcal{D}}}_F^2}{\sigma^2}} \Bigg[c_1 \left(\sigma^2\right)^{- n_L k / 2 + 1/2} + \\
				& \blankeq + \left(1 + c_2 \left(\sigma^2\right)^{- n_L k / 2} \exp{\frac{\norm{\bm{y}_{\mathcal{D}}}_F^2}{\sigma^2}} \right) c_3 \left(\sigma^2\right)^{- n_L k / 2}\Bigg] \leq \\
				& \leq c_1 \left(\sigma^2\right)^{1/2} \exp{\frac{\norm{\bm{y}_{\mathcal{D}}}_F^2}{\sigma^2}} \left(c_2 + c_3 \left(\sigma^2\right)^{1/2} + c_4 \left(\sigma^2\right)^{- n_L k / 2} \exp{\frac{\norm{\bm{y}_{\mathcal{D}}}_F^2}{\sigma^2}}\right).
			\end{aligned}
		\end{gathered}
	\end{equation*}
\end{proof}

\subsection{\texorpdfstring{1\textsuperscript{st}}{First} term bound} \label{subsec:firstterm}

As we already mentioned, in order to control the first term we need first to apply the convexity property presented in \cref{eq:wassconvex} to $\mu_{\mathrm{post}}$ and $\bar{\mu}$ defined respectively in \cref{eq:mupost,eq:barmu}.
Doing that we would have
\begin{equation} \label{eq:wassfirstterm}
	\wass[1]{\mu_{\mathrm{post}}}{\bar{\mu}} \leq \int_{\real^+} \wass[1]{\mu_{\sigma^2}(s)}{\widetilde{\mu}_{\sigma^2}(s)} \frac{\mathit{I}_{\sigma^2}(s)}{\mathit{I}} p_{\sigma^2}(s) ds
\end{equation}
where the probability measure $\nu$ in \cref{eq:wassconvex} here is simply
\begin{equation*}
	\nu \ll \lambda^+, \text{ with } \frac{d\nu}{d\lambda^+}(s) = \frac{\mathit{I}_{\sigma^2}(s)}{\mathit{I}} p_{\sigma^2}(s), \ \lambda^+ \text{ Lebesgue measure on } \real^+.
\end{equation*}
Hence, in order to get \cref{eq:wassfirstterm} we just need to prove that both $(\mu_{\sigma^2}(s))_{s \in \real^+}$ and $(\widetilde{\mu}_{\sigma^2}(s))_{s \in \real^S}$ are Markov kernels with source $\real^+$ and target $\real^{n_L \times k}$. \\
We first observe that $\forall s \in \real^+$, both $\mu_{\sigma^2}(s)$ and $\widetilde{\mu}_{\sigma^2}(s)$ are probability measures, which follows from $\mu_{\sigma^2}(s)(\real^{n_L \times k}) = \widetilde{\mu}_{\sigma^2}(s)(\real^{n_L \times k}) = 1$ and applying dominated convergence. \\
It remains only to check that for any $B \in \borel{\real^{n_L \times k}}$, $\mu_{\sigma^2}(\cdot)(B)$ and $\widetilde{\mu}_{\sigma^2}(\cdot)(B)$, are measurable from $(\real^+, \borel{\real^+})$ to $([0, 1], \borel{[0, 1]})$, which is again easy to observe:  the maps
\begin{equation*}
	s \to \int_B \mathcal{L}(\bm{z}, s) \mu(d\bm{z}), \quad s \to \int_B \mathcal{L}(\bm{z}, s) \widetilde{\mu}(d\bm{z}), \quad s \to \mathit{I}_{\sigma^2}(s), \quad s \to \widetilde{\mathit{I}}_{\sigma^2}(s) 
\end{equation*}
are continuous because Lebesgue integrals of the map $s \to \mathcal{L}(\bm{z}, s)$ with respect to $\bm{z}$, integration variable of a probability measure, and therefore we have the thesis.

\smallskip

So, the final bound on the first term can be found explicitly observing that the two probability measures $\mu_{\sigma^2}(s)$ and $\widetilde{\mu}_{\sigma^2}(s)$ parametrized by $s$, coincide with the laws of $G(\bm{x}) \given (\train, \sigma^2 = s)$ and $f_{\bm{\theta}}(\bm{x}) \given (\train, \sigma^2 = s)$. 
By \Cref{cor:convposteriorsigma}, 
\begin{equation*}
	\wass[1]{\mu_{\sigma^2}(s)}{\widetilde{\mu}_{\sigma^2}(s)} \leq h(s) \frac{c}{\sqrt{n_{min}}},
\end{equation*}
with $c$ and $h$ as in the statement of the result used, which implies
\begin{equation} \label{eq:boundfirstterm}
	\wass[1]{\mu_{\mathrm{post}}}{\bar{\mu}} \leq \frac{c}{\sqrt{n_{min}}} \int_{\real^+} h(s) \frac{\mathit{I}_{\sigma^2}(s)}{\mathit{I}} p_{\sigma^2}(s) ds.
\end{equation}
In \Cref{lem:boundexplikelihood} we already computed the bounds for $\mathit{I}_{\sigma^2}(s)$, therefore we can also bound $\mathit{I}_{\sigma^2}(s) / \mathit{I}$: observing 
\begin{equation*}
	\mathit{I}_{\sigma^2}(s) = \mean[\bm{z} \sim G(\bm{x}) | \sigma^2 = s]{\mathcal{L}(\bm{z}, s)} \quad \text{and} \quad \mathit{I} = \int_{\real^+} \mathit{I}_{\sigma^2}(s) p_{\sigma^2}(s) ds,
\end{equation*}
it is easy to check that, fixed $\epsilon < \left(\lambda_+\right)^{-1}$, with $\lambda_+ \coloneqq \max\{\lambda \given \lambda \in \operatorname{Sp}\left(\bm{K}'(\bm{x})\right)\}$, we have
\begin{align}	
	& \begin{aligned} \nonumber
		\mathit{I}_{\sigma^2}(s) & \leq c \, s^{-n_L k / 2} \exp{-\frac{1}{s} \frac{\epsilon}{2 (\epsilon + 1)} \norm{\bm{y}_{\mathcal{D}}}_F^2}, \text{ and} \\
	\end{aligned} \\
	& \begin{aligned} \label{eq:lowerboundI}
		\mathit{I} & \geq c \int_{\real^+} s^{- n_L k / 2} \exp{-\frac{\norm{\bm{y}_{\mathcal{D}}}_F^2}{s}} s^{- a - 1} \exp{-\frac{b}{s}} ds = \\
		& = c \int_{\real^+} s^{n_L k / 2} \exp{- s \norm{\bm{y}_{\mathcal{D}}}_F^2} s^{a + 1} \exp{- s b} s^{-2} ds = \\
		& = c \int_{\real^+} s^{n_L k / 2 + a - 1} \exp{- s (\norm{\bm{y}_{\mathcal{D}}}_F^2 + b)} ds = c \, \Gamma\left(n_L k /  2 + a\right).
	\end{aligned}
\end{align}
Hence, the bound in \cref{eq:boundfirstterm} can be further simplified bounding the following integral:
\begin{align*}
	& \int_{\real^+} h(s) \frac{\mathit{I}_{\sigma^2}(s)}{\mathit{I}} p_{\sigma^2}(s) ds \leq \\
	& \kern20pt \begin{aligned}
		& \leq c \int_{\real^+} \left(c_1 s^{1/2} \exp{\frac{\norm{\bm{y}_{\mathcal{D}}}_F^2}{s}} \left(c_2 + c_3 s^{1/2} + c_4 s^{- n_L k / 2} \exp{\frac{\norm{\bm{y}_{\mathcal{D}}}_F^2}{s}}\right)\right) \cdot \\
		& \blankeq \cdot s^{- n_L k / 2} \exp{-\frac{1}{s} \frac{\epsilon}{2 (\epsilon + 1)} \norm{\bm{y}_{\mathcal{D}}}_F^2 } \cdot s^{- a - 1} \exp{-\frac{b}{s}} ds = \\
		& = c_1 \int_{\real^+} s^{a + n_L k / 2 - 1/2 - 1} \exp{-s \left(b + \left(\frac{\epsilon}{2\epsilon + 2} - 1\right) \norm{\bm{y}_{\mathcal{D}}}_F^2\right)} ds \, + \\
		& \blankeq + c_2 \int_{\real^+} s^{a + n_L k / 2 - 1 - 1} \exp{-s \left(b + \left(\frac{\epsilon}{2\epsilon + 2} - 1\right) \norm{\bm{y}_{\mathcal{D}}}_F^2\right)} ds \, + \\
		& \blankeq + c_3 \int_{\real^+} s^{a + n_L k - 1/2 - 1} \exp{-s \left(b + \left(\frac{\epsilon}{2\epsilon + 2} - 2\right) \norm{\bm{y}_{\mathcal{D}}}_F^2\right)} ds = \\
		& = c_1 \Gamma\left(a + n_L k / 2 - 1/2\right) + c_2 \Gamma\left(a + n_L k / 2 - 1\right) + c_3 \Gamma\left(a + n_L k - 1/2\right),
	\end{aligned}
\end{align*}
under the assumptions 
\begin{equation*}
	a > \frac{1}{2} \quad \text{and} \quad b > \left(1 + \frac{\epsilon + 2}{2\epsilon + 2}\right) \norm{\bm{y}_{\mathcal{D}}}_F^2.
\end{equation*}
Therefore, it holds
\begin{equation*}
	\wass[1]{\mu_{\mathrm{post}}}{\bar{\mu}} \leq \frac{c}{\sqrt{n_{min}}}.
\end{equation*}

\subsection{\texorpdfstring{2\textsuperscript{nd}}{Second} term bound} \label{subsec:secondterm}

Finally, to control the second term the idea is to apply the Theorem 6.15 of \citet{otvillani2008}, which in our case is simplified to \Cref{lem:boundwasstv}. 
We use the following characterization of the total variation measure (see Section 6.1 of \citet{rcarudin1987}): $\forall A \in \borel{\real^S}$,
\begin{equation*}
	|\mu - \nu|(A) = \sup_{\left(A_i\right)_{i = 1}^{\infty}, \ \bigsqcup_{i = 1}^{\infty} A_i = A} \sum_{i = 1}^{\infty} |(\mu - \nu)(A_i)|.
\end{equation*}
Now the problem is that we do not know how to measure maps with the finite measure $|\bar{\mu} - \widetilde{\mu}_\mathrm{post}|$, but we know how to bound it. \\
Indeed, introducing the following notation to improve the readability,
\begin{equation*}
	k(s) \coloneqq \frac{\mathit{I}_{\sigma^2}(s)}{\mathit{I}} - \frac{\widetilde{\mathit{I}}_{\sigma^2}(s)}{\widetilde{\mathit{I}}},
\end{equation*}
we have
\begin{equation*}
	\bar{\mu} - \widetilde{\mu}_{\mathrm{post}} = \int_{\real^+} k(s) \widetilde{\mu}_{\sigma^2}(s) p_{\sigma^2}(s) ds,
\end{equation*}
and therefore $\forall A \in \borel{\real^{n_L \times k}}$, 
\begin{equation*}
	\begin{aligned}
		|\bar{\mu} - \widetilde{\mu}_\mathrm{post}|(A) & = \sup_{\left(A_i\right)_{i = 1}^{\infty}, \ \bigsqcup_{i = 1}^{\infty} A_i = A} \sum_{i = 1}^{\infty} \left|\int_{\real^+} k(s) \widetilde{\mu}_{\sigma^2}(s)(A_i) p_{\sigma^2}(s) ds\right| \leq \\
		& \leq \sup_{\left(A_i\right)_{i = 1}^{\infty}, \ \bigsqcup_{i = 1}^{\infty} A_i = A} \sum_{i = 1}^{\infty} \int_{\real^+} |k(s)| \widetilde{\mu}_{\sigma^2}(s)(A_i) p_{\sigma^2}(s) ds = \\
		& = \sup_{\left(A_i\right)_{i = 1}^{\infty}, \ \bigsqcup_{i = 1}^{\infty} A_i = A} \lim_{j \to \infty} \int_{\real^+} |k(s)| \widetilde{\mu}_{\sigma^2}(s)\left(\bigsqcup_{i = 1}^j A_i\right) p_{\sigma^2}(s) ds.
	\end{aligned}
\end{equation*}
It is easy to observe that it is possible to apply Dominated Convergence Theorem: $\forall j \in \natural_{> 0}$,
\begin{equation*}
	\left||k(s)| \widetilde{\mu}_{\sigma^2}(s)\left(\bigsqcup_{i = 1}^j A_i\right) p_{\sigma^2}(s)\right| \leq |k(s)| p_{\sigma^2}(s) \quad \text{and} \quad \int_{\real^+} |k(s)| p_{\sigma^2}(s) \leq 2.
\end{equation*}
Therefore,
\begin{equation*}
	\begin{aligned}
		|\bar{\mu} - \widetilde{\mu}_\mathrm{post}|(A) & \leq \sup_{\left(A_i\right)_{i = 1}^{\infty}, \ \bigsqcup_{i = 1}^{\infty} A_i = A} \int_{\real^+} |k(s)| \widetilde{\mu}_{\sigma^2}(s)\left(\bigsqcup_{i = 1}^\infty A_i\right) p_{\sigma^2}(s) ds = \\
		& = \int_{\real^+} |k(s)| \widetilde{\mu}_{\sigma^2}(s)(A) p_{\sigma^2}(s) ds \eqcolon \nu(A).
	\end{aligned}
\end{equation*}
It is easy to observe that $\nu: \borel{\real^{n_L \times k}} \to \real^+$ is a finite measure, indeed:
\begin{itemize}
	\item[-] $\nu(\emptyset) = 0$, $\nu(\real^{n_L \times k}) = \int_{\real^+} |k(s)| p_{\sigma^2}(s) ds \leq 2$;
	\item[-] given $A = \bigsqcup_{i = 1}^{\infty} A_i$, again using dominated convergence we have
		\begin{equation*}
			\begin{aligned}
				\sum_{i = 1}^{\infty} \nu(A_i) & = \sum_{i = 1}^{\infty} \int_{\real^+} |k(s)| \widetilde{\mu}_{\sigma^2}(A_i) p_{\sigma^2}(s) ds = \int_{\real^+} |k(s)| \widetilde{\mu}_{\sigma^2}\left(\bigsqcup_{i = 1}^{\infty} A_i\right) p_{\sigma^2}(s) ds = \\
				& = \int_{\real^+} |k(s)| \widetilde{\mu}_{\sigma^2}(A) p_{\sigma^2}(s) ds = \nu(A).
			\end{aligned}
		\end{equation*}
\end{itemize}
Applying \Cref{lem:boundwasstv} and observing that if $\forall A \in \borel{\real^{n_L \times k}}$, $|\bar{\mu} - \widetilde{\mu}_\mathrm{post}|(A) \leq \nu(A)$ then $\forall f$ measurable from $\real^{n_L \times k}$ to $\real^+$, $|\bar{\mu} - \widetilde{\mu}_\mathrm{post}|(f) \leq \nu(f)$, we get
\begin{align} 
	\wass[1]{\bar{\mu}}{\widetilde{\mu}_{\mathrm{post}}} & \leq \int_{\real^{n_L \times k}} \norm{\bm{z}}_F d |\bar{\mu} - \widetilde{\mu}_\mathrm{post}|(\bm{z}) \leq \int_{\real^{n_L \times k}} \norm{\bm{z}}_F d \nu(\bm{z}) = \nonumber \\
	& = \int_{\real^+} |k(s)| p_{\sigma^2}(s) \int_{\real^{n_L \times k}} \norm{\bm{z}}_F \widetilde{\mu}_{\sigma^2}(s)(d \bm{z}) ds = \nonumber \\
	& = \int_{\real^+} |k(s)| p_{\sigma^2}(s) \frac{1}{\widetilde{\mathit{I}}_{\sigma^2}(s)} \int_{\real^{n_L \times k}} \norm{\bm{z}}_F \mathcal{L}(\bm{z}, s) \widetilde{\mu}(d\bm{z}) ds, \label{eq:wassbound1}
\end{align}
where the identity from the 1\textsuperscript{st} to the 2\textsuperscript{nd} line follows by Fubini's Theorem. \\
The inner integral is easy to compute bounding the likelihood $\mathcal{L}$ in terms of the variable $\bm{z}$, already computed in \Cref{prop:likelihood}: $\norm{\mathcal{L}}_{\infty} = c \, s^{-n_L k / 2}$. Using this result we obtain
\begin{align*}
	\int_{\real^{n_L \times k}} \norm{\bm{z}}_F \mathcal{L}(\bm{z}, s) \widetilde{\mu}(d\bm{z}) & = \mean[\bm{z} \sim f_{\bm{\theta}}(\bm{x}) \given \sigma^2 = s]{\norm{\bm{z}}_F \mathcal{L}(\bm{z}, s)} \leq c \, s^{-n_L k / 2} \mean[\bm{z} \sim f_{\bm{\theta}}(\bm{x}) \given \sigma^2 = s]{\norm{\bm{z}}_F}.
\end{align*}
Now the procedure to compute the first moment of the distribution of $ f_{\bm{\theta}}(\bm{x}) \given \sigma^2 = s$ is analogous to the one used in the proof of \Cref{lem:boundexplikelihood} for the lower bound related to $f_{\bm{\theta}}(\bm{x})$.
First we recall that $f_{\bm{\theta}'} = s^{-1/2} f_{\bm{\theta} \given \sigma^2 = s}$ and also $G' = s^{-1/2} (G \given \sigma^2 = s)$.
Then we apply the change of variable $\bm{u} = \bm{z} / s^{1/2}$, and we get
\begin{equation*}
	\begin{aligned}
		\mean[\bm{z} \sim f_{\bm{\theta}}(\bm{x}) \given \sigma^2 = s]{\norm{\bm{z}}_F} & = \int_{\real^{n_L \times k}} \norm{\bm{z}}_F d \mathbb{P}_{f_{\bm{\theta}}(\bm{x}) \given \sigma^2 = s}(\bm{z}) = \\
		& = \int_{\real^{n_L \times k}} \norm{\bm{u} s^{1/2}}_F s^{n_L k / 2} d \mathbb{P}_{f_{\bm{\theta}'}(\bm{x})}(\bm{\bm{u}}) = \\
		& = s^{n_L k / 2 + 1/2} \mean[\bm{u} \sim f_{\bm{\theta}'}(\bm{x})]{\norm{\bm{u}}_F}.
	\end{aligned}
\end{equation*}
Now that we removed the dependence in terms of $s$ it is possible to bound the moment of the rescaled BNN using the moment of the rescaled NNGP, which can be upper bounded as in \cref{eq:p3-4}:
\begin{equation*}
	\mean[\bm{u} \sim G'(\bm{x})]{\norm{\bm{u}}_F} \leq (n_L k)^{\frac{1}{2}} \cdot \left(\max_{i \in [k]} \mean{\norm{\varphi_L\left(G^{(L - 1)}(\bm{x}_i)\right)}_2^2} / n_L + 1\right)^{\frac{1}{2}},
\end{equation*}
with a right-hand side that is just a constant term depending on $\bm{x}, \bm{\varphi}, (\bm{\sigma}_l)_{l = 1}^{L - 1}, n_L$. \\
In order to do so we replicate an analogue of \cref{eq:boundmeanexpscaled}: thanks to the triangle inequality we know $\lipschitz{\norm{\bm{\cdot}}_F} = 1$, and applying \Cref{thm:priornngp} we get
\begin{equation*}
	\left|\mean[\bm{u} \sim G'(\bm{x})]{\norm{\bm{u}}_F} - \mean[\bm{u} \sim f_{\bm{\theta}}'(\bm{x})]{\norm{\bm{u}}_F}\right| \leq \lipschitz{\norm{\bm{\cdot}}_F} \wass[1]{G'(\bm{x})}{f_{\bm{\theta}'}(\bm{x})} \leq \frac{c}{\sqrt{n_{min}}}.
\end{equation*}
Finally, assuming $\sqrt{n_{min}} \geq 2 c / \mean[\bm{u} \sim G'(\bm{x})]{\norm{\bm{u}}_F}$, we derive
\begin{equation*}
	\mean[\bm{u} \sim f_{\bm{\theta}}'(\bm{x})]{\norm{\bm{u}}_F} \leq \frac{3}{2} \mean[\bm{u} \sim G'(\bm{x})]{\norm{\bm{u}}_F},
\end{equation*}
which implies 
\begin{equation*}
	\int_{\real^{n_L \times k}} \norm{\bm{z}}_F \mathcal{L}(\bm{z}, s) \widetilde{\mu}(d\bm{z}) \leq c s^{1/2}.
\end{equation*}
Restarting from the result in \cref{eq:wassbound1} and using the lower bound related to $f_{\bm{\theta}}(\bm{x})$ in \Cref{lem:boundexplikelihood} we get
\begin{align*}
	\wass[1]{\bar{\mu}}{\widetilde{\mu}_{\mathrm{post}}} & \leq c \int_{\real^+} |k(s)| \frac{s^{1/2}}{\widetilde{\mathit{I}}_{\sigma^2}(s)} p_{\sigma^2}(s) ds \leq \\
	& \leq c \int_{\real^+} |k(s)| \exp{\frac{\norm{\bm{y}_{\mathcal{D}}}_F^2}{s}} s^{1/2} s^{-a - 1} \exp{-\frac{b}{s}} ds. 
\end{align*}
Now it is sufficient to show 
\begin{equation} \label{eq:boundks}
	|k(s)| = \left|\frac{\mathit{I}_{\sigma^2}(s)}{\mathit{I}} - \frac{\widetilde{\mathit{I}}_{\sigma^2}(s)}{\widetilde{\mathit{I}}}\right| \leq s^{-n_L k / 2} \frac{c}{\sqrt{n_{min}}}
\end{equation}
to have the thesis. Indeed, we would have
\begin{align*}
	\wass[1]{\bar{\mu}}{\widetilde{\mu}_{\mathrm{post}}} & \leq \frac{c}{\sqrt{n_{min}}} \int_{\real^+} s^{-n_L k / 2 - a + 1/2 - 1} \exp{-\frac{1}{s} \left(b - \norm{\bm{y}_{\mathcal{D}}}_F^2\right)} ds = \nonumber \\
	& = \frac{c}{\sqrt{n_{min}}} \Gamma\left(n_L k / 2 - 1/2 + a\right) = \frac{c}{\sqrt{n_{min}}}, 
\end{align*}
under the assumption $b > \norm{\bm{y}_{\mathcal{D}}}_F^2$. \\
Let us thus prove \cref{eq:boundks}. We begin by observing that we already know how to bound the absolute difference $\big|\mathit{I}_{\sigma^2}(s) - \widetilde{\mathit{I}}_{\sigma^2}(s)\big|$, using the same arguments applied in the proof of \Cref{lem:boundexplikelihood}:
\begin{equation} \label{eq:absdiffIstildeIs}
	\begin{aligned}
		\left|\mathit{I}_{\sigma^2}(s) - \widetilde{\mathit{I}}_{\sigma^2}(s)\right| & \leq \lipschitz{\mathcal{L}(\cdot, s)} \wass[1]{G(\bm{x})}{f_{\bm{\theta}}(\bm{x})} \leq c s^{-n_L k / 2 - 1/2} \cdot s^{1/2} \frac{c}{\sqrt{n_{min}}} \leq \\
		& \leq s^{- n_L k /2} \frac{c}{\sqrt{n_{min}}}.
	\end{aligned}
\end{equation}
Therefore, it remains to show that $\mathit{I}^{-1} \leq c$ and $\widetilde{\mathit{I}}^{-1} \geq c$ for some $c$ depending only on the usual parameters.
Indeed, it would yield
\begin{equation*}
	\left|\frac{\mathit{I}_{\sigma^2}(s)}{\mathit{I}} - \frac{\widetilde{\mathit{I}}_{\sigma^2}(s)}{\widetilde{\mathit{I}}}\right| \leq c \left|\mathit{I}_{\sigma^2}(s) - \widetilde{\mathit{I}}_{\sigma^2}(s)\right| \leq s^{- n_L k /2} \frac{c}{\sqrt{n_{min}}}.
\end{equation*}
We already saw $\mathit{I} \geq c$ in \cref{eq:lowerboundI}, but in order to find an upper bound for $\widetilde{\mathit{I}}$ we also need an upper bound for $\mathit{I}$. 
The procedure to get it is analogous to the one used in \cref{eq:lowerboundI}: starting from \Cref{lem:boundexplikelihood} we bound the negative exponential with $1$, and we are done,
\begin{align*}
	\mathit{I} & \leq c \int_{\real^+} s^{-n_L k / 2} \exp{-\frac{1}{s} \frac{\epsilon}{2 (\epsilon + 1)} \norm{\bm{y}_{\mathcal{D}}}_F^2} p_{\sigma^2}(s) ds \leq c \int_{\real^+} s^{-n_L k / 2 - a - 1} \exp{-\frac{b}{s}} ds = \\
	& = c \Gamma\left(n_L k / 2 + a\right).
\end{align*}
To prove $\widetilde{\mathit{I}} \leq c$ it is sufficient to observe that 
\begin{align*}
	\left|\mathit{I} - \widetilde{\mathit{I}}\right| & = \left|\int_{\real^+} \left(\mathit{I}_{\sigma^2}(s) - \widetilde{\mathit{I}}_{\sigma^2}(s)\right) p_{\sigma^2}(s) ds\right| \leq \int_{\real^+} \left|\mathit{I}_{\sigma^2}(s) - \widetilde{\mathit{I}}_{\sigma^2}(s)\right| p_{\sigma^2}(s) ds \leq \\
	& \leq \frac{c}{\sqrt{n_{min}}} \int_{\real^+} s^{-n_L k / 2 - a - 1} \exp{-\frac{b}{s}} ds = \frac{c}{\sqrt{n_{min}}},
\end{align*}
where from the 1\textsuperscript{st} to the 2\textsuperscript{nd} line we used the inequality in \cref{eq:absdiffIstildeIs}.
Hence, considering $n_{min}$ sufficiently large to have $\sqrt{n_{min}} > 2c / \mathit{I}$, it follows $|\mathit{I} - \widetilde{\mathit{I}}| \leq \mathit{I} / 2$ which implies $\widetilde{\mathit{I}} \leq 3/2 \, \mathit{I} \leq c$.

\section{Simulations details} \label{sec:simulationsdetails}

Starting from the hierarchical model in \cref{eq:bnnhiermodel}, to sample from the posterior BNN using \Cref{alg:samplingpostbnn}, we only need to explicitly define a method for sampling from $\sigma^2 \given \bm{\theta}, \train$. 
To achieve this, it is sufficient to retrieve the kernel of its density, which can be written explicitly. 
Recalling that $\bm{x}_{\mathcal{D}}$ is assumed independent of $\bm{\theta}$ and is also obviously independent of $\sigma^2$, it holds
\begin{equation} \label{eq:conditionaldistsigmaprod}
	\begin{aligned}
		p_{\sigma^2 | \bm{\theta}, \bm{x}_{\mathcal{D}}, \bm{y}_{\mathcal{D}}}(\sigma^2) & = \frac{p_{\sigma^2, \bm{\theta}, \bm{x}_{\mathcal{D}}, \bm{y}_{\mathcal{D}}}(\sigma^2, \bm{\theta}, \bm{y}_{\mathcal{D}}, \bm{x}_{\mathcal{D}})}{p_{\bm{\theta}, \bm{x}_{\mathcal{D}}, \bm{y}_{\mathcal{D}}}(\bm{\theta}, \bm{x}_{\mathcal{D}}, \bm{y}_{\mathcal{D}})} \propto p_{\sigma^2}(\sigma^2) \, p_{\bm{\theta} \given \sigma^2}(\bm{\theta}) \, p_{\bm{x}_{\mathcal{D}}}(\bm{x}_{\mathcal{D}}) \, p_{\bm{y}_{\mathcal{D}} | \bm{\theta}, \bm{x}_{\mathcal{D}}, \sigma^2}(\bm{y}_{\mathcal{D}}) \propto \\
		& \propto p_{\sigma^2}(\sigma^2) \, p_{\bm{W}^{(L)} \given \sigma^2}\left(\bm{W}^{(L)}\right) \, p_{\bm{b}^{(L)} \given \sigma^2}\left(\bm{b}^{(L)}\right) \,  p_{\bm{y}_{\mathcal{D}} | \bm{\theta}, \bm{x}_{\mathcal{D}}, \sigma^2}(\bm{y}_{\mathcal{D}}).
	\end{aligned}
\end{equation}
Exploiting
\begin{equation*}
	f_{\bm{\theta}}(\bm{x}_{\mathcal{D}}) = \left(\sigma^2\right)^{1/2} \left(\frac{\bm{W}^{(L)}}{\left(\sigma^2\right)^{1/2}} \varphi\left(f^{(L - 1)}_{\bm{\theta}}(\bm{x}_{\mathcal{D}})\right) + \frac{\bm{b}^{(L)}}{\left(\sigma^2\right)^{1/2}}\right) \eqcolon \left(\sigma^2\right)^{1/2} f_{\bm{\theta}'}(\bm{x}_{\mathcal{D}}),
\end{equation*}
with $f_{\bm{\theta}'}(\bm{x}_{\mathcal{D}})$ independent of $\sigma^2$ (as in \cref{eq:thetaprime}), we have also
\begin{equation} \label{eq:densities}
	\begin{aligned}
		p_{\sigma^2}(\sigma^2) & \propto \left(\sigma^2\right)^{-(a + 1)} \exp{-\frac{b}{\sigma^2}}, \\
		p_{\bm{W}^{(L)} \given \sigma^2}(\bm{W}^{(L)}) & \propto \left(\sigma^2\right)^{-n_{L} n_{L - 1} / 2} \exp{-\frac{n_{L - 1}}{2 \sigma^2} \norm{\bm{W}^{(L)}}^2_F}, \\
		p_{\bm{b}^{(L)} \given \sigma^2}(\bm{b}^{(L)}) & \propto \left(\sigma^2\right)^{-n_{L}/2} \exp{-\frac{1}{2 \sigma^2} \norm{\bm{b}^{(L)}}^2_F}, \\
		p_{\bm{y}_{\mathcal{D}} | \bm{\theta}, \sigma^2}(\bm{y}_{\mathcal{D}}) & \propto \left(\sigma^2\right)^{-n_L k/ 2} \exp{-\frac{1}{2\sigma^2} \norm{\bm{y}_{\mathcal{D}} - f_{\bm{\theta}}(\bm{x}_{\mathcal{D}})}_F^2} = \\
		& = \left(\sigma^2\right)^{-n_L k / 2} \exp{-\frac{1}{2 \sigma^2} \norm{\bm{y}_{\mathcal{D}}}_F^2} \cdot \exp{-\frac{1}{2} \norm{f_{\bm{\theta}'}(\bm{x}_{\mathcal{D}})}_F^2} \cdot \\
		& \blankeq \cdot \exp{+\frac{1}{\left(\sigma^2\right)^{1/2}} \flatten{\bm{y}_{\mathcal{D}}}^T \flatten{f_{\bm{\theta}'}(\bm{x}_{\mathcal{D}})}}.
	\end{aligned}
\end{equation}
Hence, substituting the identities reported in \cref{eq:densities} inside \cref{eq:conditionaldistsigmaprod} we get
\begin{equation} \label{eq:conditionaldistsigma}
	\begin{aligned}
		p_{\sigma^2 | \bm{\theta}, \train}(\sigma^2) \propto \left(\sigma^2\right)^{-(a' + 1)} \exp{-\frac{b'}{\sigma^2}} \exp{+\frac{c'}{\left(\sigma^2\right)^{1/2}}},
	\end{aligned}
\end{equation}
with $a', b' \in \real^+$, $c' \in \real$ such that
\begin{equation*}
	\begin{aligned}
		a' & \coloneqq a + (n_{L - 1} + k + 1) n_L / 2, \\
		b' & \coloneqq b + \frac{1}{2} \left(n_{L - 1} \norm{\bm{W}^{(L)}}^2_F + \norm{\bm{b}^{(L)}}^2_F + \norm{\bm{y}_{\mathcal{D}}}_F^2\right), \\
		c' & \coloneqq \flatten{\bm{y}_{\mathcal{D}}}^T \flatten{f_{\bm{\theta}'}(\bm{x}_{\mathcal{D}})}.
	\end{aligned}
\end{equation*}
As mentioned in \Cref{sec:simulations}, we have used the companion library of \citet{ding2022} to sample from the density in \cref{eq:conditionaldistsigma} in the \texttt{Python} implementation of \Cref{alg:samplingpostbnn}. This has been possible because, given $x$ random variable with density
\begin{equation*}
	p_{x}(x) \propto x^{-(a + 1)} \exp{-\frac{b}{x}} \exp{\frac{c}{x^{1/2}}} \indic{\{x \geq 0\}},
\end{equation*}
the normalization constant is
\begin{align*}
	\int_{0}^{+\infty} p_{x}(x) &= \int_{0}^{+\infty} x^{-(a + 1)} \exp{-\frac{b}{x}} \exp{\frac{c}{x^{1/2}}} dx = \\
	& = 2 \int_{0}^{+\infty} y^{2a + 1} \exp{-b y^2} \exp{cy} dy = \\
	&= 2 (2b)^{-a} \Gamma(2a) \exp{\frac{c^2}{8b}} D_{-2a}\left(\frac{-c}{(2b)^{1/2}}\right),
\end{align*}
with $D$ being the parabolic cylinder function. Therefore, given $z \sim \mathcal{GIN}^+(2a + 1, c/2b, \sqrt{1/2b})$ (as defined by \citet[Appendix D]{ding2022}), applying the transformation $y \coloneqq x^2$, we get a random variable with the same distribution as our target $x$: $x \sim y$.  

\begin{remark}
	The simulations closely follow the theoretical framework developed in this work. 
	However, during the sampling of the posterior Student-$t$ process and BNNs, it is performed a rescaling of $\sigma^2$. This adjustment is applied where $\sigma^2$ is used as the variance of $\bm{y}_{\mathcal{D}} \given \bm{\theta}, \bm{x}_{\mathcal{D}}, \sigma^2$, in order to address numerical stability issues encountered during the sampling process described in \Cref{alg:samplingpostbnn}.
\end{remark}

\end{document}